\documentclass[mnsc,nonblindrev]{informs3} 

\def\V{{\mathbb{V}}\,}
\def\E{{\mathbb{E}}\,}
\def\Cov{{\mathrm{Cov}}\,}

\usepackage{times}
\usepackage{soul}
\usepackage{url}
\usepackage[hidelinks]{hyperref}
\usepackage[utf8]{inputenc}
\usepackage[small]{caption}
\usepackage{graphicx}
\usepackage{amsmath}
\usepackage{booktabs}
\usepackage{xcolor}
\usepackage{comment}
\usepackage{amsfonts}
\usepackage{bm}
\usepackage{booktabs}
\usepackage{subfig}

\usepackage{natbib}
 \bibpunct[, ]{(}{)}{,}{a}{}{,}%
 \def\bibfont{\small}%
\TheoremsNumberedThrough     
\ECRepeatTheorems

\EquationsNumberedThrough    

\MANUSCRIPTNO{}

\graphicspath{{figures/}}
\makeatletter
\def\input@path{{sections/}}
\makeatother

\begin{document}


\RUNAUTHOR{Nabi et al.} 

\RUNTITLE{Bayesian Meta-Prior Learning Using Empirical Bayes}

\TITLE{Bayesian Meta-Prior Learning \\ Using Empirical Bayes}

\ARTICLEAUTHORS{
 \AUTHOR{Sareh Nabi}
  \AFF{Foster School of Business, University of Washington, Seattle, Washington 98195, 
  \EMAIL{snabi@uw.edu}} 
 \AUTHOR{Houssam Nassif, Joseph Hong}
  \AFF{Amazon, Seattle, Washington 98109, \EMAIL{houssamn@amazon.com}, \EMAIL{josehong@amazon.com}} 
 \AUTHOR{Hamed Mamani}
  \AFF{Foster School of Business, University of Washington, Seattle, Washington 98195, 
	\EMAIL{hmamani@uw.edu}}
	 \AUTHOR{Guido Imbens}
  \AFF{Graduate School of Business, Stanford University, California 94305,  
	\EMAIL{gimbens@amazon.com}}
} 

\ABSTRACT{
Adding domain knowledge to a learning system is known to improve results. 
In multi-parameter Bayesian frameworks, such knowledge is incorporated as a 
prior. On the other hand, the various model parameters 
can have different learning rates in real-world problems, especially with skewed data.
Two often-faced challenges in Operation Management and Management Science applications 
are the absence of informative priors, and the inability to control parameter learning rates.
In this study, we propose a hierarchical Empirical Bayes approach that addresses both challenges,
and that can generalize to any Bayesian framework.
Our method learns empirical meta-priors from the data itself, 
and uses them to decouple the learning rates of first-order and 
second-order features (or any other given feature grouping) in a Generalized Linear Model. 
As the first-order features are 
likely to have a more pronounced effect on the outcome, focusing on learning first-order 
weights first is likely to improve performance and convergence time. Our Empirical Bayes 
method clamps features in each group together and uses the deployed model's observed data 
to empirically compute a hierarchical prior in hindsight. 
We report theoretical results for the unbiasedness, strong consistency, and optimal frequentist cumulative regret properties of our meta-prior variance estimator.
We apply our method to a standard 
supervised learning optimization problem, as well as an online combinatorial optimization problem
in a contextual bandit setting implemented in an Amazon production system. Both during 
simulations and live experiments, our method shows marked improvements, especially in cases 
of small traffic. Our findings are promising, as optimizing over sparse data is often a 
challenge. 
}



\maketitle

\newpage

\section{Introduction} 

We start by introducing and motivating the problem at hand.

\subsection{Sequential Decisions and Prior Knowledge}
Many scenarios arise in real-world applications where a decision-making agent 
must interact with an unknown environment in a sequential fashion, 
in which the agent receives a reward signal after each selected 
action~\citep{Ch13, SuBa18, Naka18, biswas2019seeker, SuBi19, pqrSinong20}. 
The agent's objective is to find the best action-selection strategy that 
maximizes its long-run cumulative reward while navigating the existent uncertainty. 

One application of such scenarios is dynamic pricing,
where prices are dynamically adjusted in 
response to real-time demand and supply information. 
Nowadays dynamic pricing is practiced in e-commerce, travel
industries, ride-hailing apps, smart retail electricity pricing, 
to name a few \citep{KaLo13, BaKe17, FeSi18, LeJa18, KeLiSu20}. 
Firms often face uncertainties such as unknown demand and 
customer preferences when making pricing decisions. Their challenge is to 
learn these uncertainties to better optimize revenue.  

Online advertising is another example of sequential decision making 
under uncertainty \citep{nassif2016music, ChSi17, SawantHVAE18, Ra19, RaYo21}. 
According to a recent Interactive Advertising Bureau
(IAB) report, the digital ad revenue in the US hit a record of $\$49.5$ 
billions in the first half of 2018, an increase of 
$23\%$ compared to the same period last year \citep{IAB}. 
In search advertising, an agent (like a search engine)
makes decisions sequentially on what ads to show when a user renders a webpage.
As the search engine's revenue depends on whether the user clicks on the ad, 
the agent's objective is to maximize revenue by accurately predicting the 
click-through-rate. Obtaining such predictions is a challenge, specially 
in cold start settings where the search engine has little or 
no knowledge of the user.

A main characteristic that arises in these aforementioned scenarios 
is the trade-off between the exploration involved in 
learning the unknown environment and the exploitation which occurs due
to reward maximization. One way of solving this explore/exploit trade-off 
is by using the Multi-Armed Bandit (MAB) approach, originally proposed
by~\citet{Ro52}. Since then many studies have worked on MAB modeling 
framework and proposed algorithms to address the explore/exploit trade-off.  

This work focuses on Bayesian MABs, which allow the encoding of
prior knowledge into the learning process and thereby increase
its efficiency \citep{GhMa15, LaSz20}. Most such algorithms are based on 
Thompson Sampling (TS) \citep{Th33}. The TS procedure 
selects an action proportionally to its probability of
being optimal, conditioned on previous observations. 
This algorithm has demonstrated promising empirical results 
and is widely applied in industry and academia.

The choice of an appropriate informative prior in a Bayesian framework is often challenging, 
especially for reward models with multiple unknown parameters as in contextual MABs. 
This is why Bayesian MABs often assume a non-informative 
prior. Recent literature demonstrates that the design of a prior can 
have significant impact on the performance of Bayesian bandits \citep{BaSi19}. 
Can we construct a general framework that computes an empirical prior from early
randomized data? Given a categorization of features in a model, can we decouple
the learning rates of the parameters in each category? To the best of our knowledge, 
there is no work in the literature that addresses these questions broadly.
This study aims at providing a general solution in estimating empirical 
priors for a wide range of applications that are modeled as Bayesian 
bandits or involve Bayesian learning.

\subsection{Research Agenda and Our Approach}
Our objective in this study is to utilize early randomized data 
to construct an experiment-specific hierarchical informative prior that 
can be applied to any individual problem instance modeled as
a Bayesian framework. Such hierarchical informative priors
can decouple the parameter learning rates, improve MAB optimization,
 and potentially lead to higher 
cumulative reward and shorter convergence time. 

We are motivated by our production setting in Amazon, where
we aim at optimizing a webpage layout with multiple components \citep{hill2017kdd}. 
In the simplest case, we have one feature per
component value (e.g. ``image2''), and one interaction feature between each pair of components
(e.g. ``image2 AND title1''). We refer to the first type of features as ``first-order features'',
and to the second type as ``second-order features''. These two features types are 
functionally distinct.  Based on analysis of fully-converged past experiments, 
we observe that the second-order feature weights are clustered more closely around the mean,
while the first-order feature weights have larger absolute values, with a more pronounced effect on the outcome. 
Mathematically, this translates into a wider variance for the estimated first-order weights, when compared to the second-order ones. Can we leverage such meta-variance to learn the first-order order 
weights first, then move to the second-order weights as more data becomes available?

In our use-case, this question is particularly prescient, as the first and second order features have different incidence rates. Let $n$ be the number of values per component (e.g., $n=4$ possible images). As time
progresses, we observe the first-order feature values at a rate of $1/n$, while we observe the second-order
feature values at a slower rate of $1/n^2$. Although our proposed method does not require 
a different incidence rate, let alone a higher rate for the higher importance category, this research was partly motivated by this observation.

We leverage Empirical Bayes (EB) techniques to extract better 
priors from early randomized data. Empirical Bayes,
also known as maximum marginal likelihood, allows for hyper-parameter 
estimation at the highest level of the hierarchical Bayes models. 
These point estimates can be obtained using either parametric or
non-parametric approaches. In essence, EB is a statistical inference
procedure where the prior distribution is estimated empirically 
(and frequentistically) from the data. It exploits the
finding that large datasets of parallel situations
carry within them their own Bayesian information \citep{EfHa16}.

Our approach applies EB to Bayesian bandits to compute 
an informative prior in hindsight, and then use this prior
to improve cumulative reward and convergence time.
We use the early random (or pseudo-random) MAB traffic to compute
the empirical prior. We then rewind the bandit, re-training it on the same traffic,
augmented with the empirical prior. 
By grouping the first-order and second-order features together, imposing a hierarchical prior, and estimating the priors by applying the EB framework, we are able to decouple the learning rates of these 
two categories of features (or any other arbitrary feature grouping). 
Note that although we have a bandit use-case, 
our method can also be applied to Bayesian learning in a standard classification setting,
as we also show. 
 
\subsection{Contributions and Findings}
Our main contribution in this study are three-folded:

First, we derive a general formula to compute a hierarchical meta-prior variance for any Bayesian 
framework using empirical Bayes. We then provide an estimator to estimate 
the EB prior variance empirically using early randomized data. Our meta-prior estimator 
derivation is general and applicable to a wide range of applications that are modeled as Bayesian 
learners. Our framework enables decoupling parameter learning 
rates by categorizing features arbitrarily, imposing a hierarchical prior per each category,
and estimating the meta-priors using our EB approach.
 
Second, we report theoretical results for the unbiasedness, consistency, and regret properties of our 
meta-prior variance estimator. We provide a closed form for the expectation of our 
estimator and articulate the amount of bias in terms of the variances of feature
estimates. We further show that our estimator is unbiased in our use-cases, which 
involve a Generalized Linear Model (GLM) used as a supervised classifier and as a bandit. We prove the strong consistency of our meta-prior variance estimator in the regime where the feature estimates are independent.
Furthermore, we show that an EB TS algorithm retains an optimal $\tilde{O}(d^{3/2} \sqrt{T})$ regret bound.

Third, we validate our meta-prior variance estimator and test its generalization by 
performing simulations on the publicly available Adult UCI Machine Learning Repository 
dataset \citep{BlMe98}. The task is to predict whether one's income 
exceeds a certain amount per year based on census data. We study
the effects of first-order features, small batch data,
prior reset time, and prior variance. We further examine the performance
of empirical Bayes prior on our Amazon live production system,
using a TS generalized linear multi-armed bandit with a probit link function
 \citep{hill2017kdd}. 
Specifically, we study the selection of a web page layout with the objective of maximizing
the likelihood of a service purchase. 
Both during simulations and live experiments, we observe marked 
improvements when empirical Bayes prior is used instead of a 
non-informative prior. We observe a more pronounced improvement in small 
traffic situations. 

The rest of this paper is structured as follows. In 
Section~\ref{sec:literature}, we review the related literature. 
In Section~\ref{EB}, we derive our empirical Bayes prior estimation and provide statistical properties 
and regret bounds in Section~\ref{sec:theory}. 
Pre-processing steps and simulation experiments using a real-world supervised dataset
are explained in Section~\ref{sec:simulationExp}.
Section~\ref{sec:liveExpAll} reports results on bandit live experiments in an Amazon production setting. 
We discuss our findings and conclude in Section~\ref{Discussion}. 

\section{Related Literature} \label{sec:literature}

In applications of sequential decision making under uncertainty, 
we face an inherent trade-off between the exploration involved in learning
unknown parameters, and the exploitation which occurs due to reward optimization. 
One way of modeling this trade-off is by using a Multi-Armed Bandit (MAB), originally proposed by \citet{Ro52}.
Many algorithms have been proposed to solve stochastic MAB problems and 
generalized linear bandits \citep{Gi89,AuCe02,GaCa11,BuCe12, BuCeKa12, 
Agrawal2013ContextualTSBounds, cesa2017boltzmann, riquelme2018deep, biswas2019seeker, DiHs20, jun2021improved}.

Our focus in this work is on Bayesian MABs that often assume 
non-informative priors \citep{KaCa12,AgGo13a,SawantHVAE18}.
Most such algorithms are based on Thompson Sampling, with promising empirical results across industry \citep{Sc10, ChLi11, TaRo13, Teo2016airstream}.
Multiple researchers investigated its theoretical guarantees. \citet{KaKo12} 
and \citet{AgGo13a} derived the asymptotic optimality of TS using non-informative priors 
and Bernoulli reward models. \citet{KoKa13} extended the work to one-dimensional
exponential family bandits and derived an asymptotic optimality with Jeffreys prior. 
Refer to~\citet{RuVa18} for a tutorial on TS algorithm.

Recent research reveals that incorporating prior knowledge
in Baysian MABs improves bandit performance and increase its efficiency 
\citep{GhMa15, LaSz20, BaBa20}. 
\citet{RaNg06} propose a transfer learning algorithm that constructs 
a multivariate Gaussian prior for a supervised learning task.
\citet{HoTa14} showed that TS is asymptotically optimal using a uniform 
prior when the reward model is Gaussian with unknown means and variances. 
But they prove that TS with the choice of Jeffreys prior and reference prior 
does not lead to asymptotic optimality. They report that choice of priors can play a
crucial role in attaining the asymptotic optimality when the reward model
has more than one unknown parameter. 

Note that our prior in question is the Bayesian prior, a probability distribution that represents one's beliefs before data is encountered. One can also encode previous domain knowledge using first-order logic rules, as in Inductive Logic Programming and Statistical Relational Learning~\citep{RDP, UM-ROC, getoor2007introduction}. Such a prior knowledge also improves bandit and classifier performance~\citep{Upgrades, Upgrades2, ILPBandit21}, but is outside the scope of this research.

Bayesian (or expected) regret can be used as a performance measure in Bayesian MABs.
By assuming an arbitrary prior distribution on reward models, 
Bayesian regret represents the average frequentist regret where the expectation 
is taken with respect to the prior on reward models. 
\citet{RuVa14, RuVa16}, and \citet{BuLi13} quantified the Bayesian regret
of TS algorithm with an informative prior.  
\citet{LiLi16} also studied the sensitivity of TS performance over the choice of priors
with the focus on the frequentist regret. They obtained regret bounds that depend
on the probability mass a given prior places on the true unknown reward model.  

Empirical Bayes (EB) is a known methodology to construct 
informative priors from early randomized data \citep{Robbins55, Efron72,Morris83,Berger85}. 
For comprehensive reviews on EB methods, refer to \citet{CaLo10,Ef12}, and \citet{Ma18}. 
We implement our prior construction framework on a TS-based GLM bandit model with 
probit link function and achieve optimal asymptotic frequentist regret 
\citep{AbLa17, hamidi2020worstcase}. 

The work of \citet{BaSi19} is relevant to our study, as they propose a transfer learning algorithm that learns the meta-prior across experiments for similar products in a dynamic pricing setting. This study assumes the existence of a sequence of related experiments, where one leverages earlier bandit experiments to compute a meta-prior to be applied to later bandit experiments. \citet{BaSi19} demonstrate empirically that leveraging the prior knowledge in Thompson Sampling, instead of following a prior-independent approach, improves performance. The approach of \citet{BaSi19} only applies in transfer learning cases, where one is performing a sequence of related experiments. For example, in a dynamic pricing setting with unknown demand, if similar products are being sold, the transfer learning technique can help expedite the convergence to the optimal price for the later products in the experiments. 

On the other hand, we do not attempt any transfer learning between experiments but rather bootstrap the early phase of an experiment to compute an empirical meta-prior for that experiment. To continue the example above, our framework aims at speeding up convergence to the optimal price using only the experiment’s data, and hence can be applied in the absence of a proper sequence of similar experiments. To adopt our approach for the pricing experiments in \cite{BaSi19}, one has to run each pricing experiment separately, collect all the data across experiments, and apply our prior formulation on the aggregated data to estimate the prior. One could potentially combine both approaches in transfer learning settings, where we compute a transfer learning prior across the sequence of similar experiments and use it to seed the empirical prior computation for each individual experiment. We leave such an extension for future work.


There exists a rich literature in computer science on meta-learning, where one solves a novel task efficiently by using prior experience on similar tasks. Often assuming a hierarchical data structure, the objective in 
meta-learning problems is to train a model that generalizes across tasks 
\citep{Sch87, Be90, Na92, Th12}. Meta-learning problems are often formulated
using a hierarchical Bayes model \citep{Go80, Be13} or an empirical Bayes model 
\citep{Ro92, Gr18, Hu20}. Bayesian meta-learning helps with fast and robust adaptation to new tasks due to 
its probabilistic nature in measuring uncertainties \citep{RaBe19, YoKi18, NiAc18, 
GoBr18}. More recent work shows gradient-based meta-learning 
approaches to be effective in few-shot learning \citep{FiAb17, FiXu18, RaFi19, ZoLu20}. 

Our meta-prior framework can potentially be applied
to dynamic Bayesian learning settings with incomplete information.
One such application is Bayesian dynamic pricing (e.g., \cite{HaKe12, ChWa16}). 
\cite{BiFe20} and \cite{KaKe2020} study this problem in the context of 
spread betting markets and service subscriptions, respectively.
\cite{KeLiSu20} study dynamic pricing problems with demand learning in the presence
of customers with heterogeneous valuations for quality. 
Another avenue of potential application is optimal stopping 
problems in a regime of dynamic learning with incomplete information. 
\cite{HaSu15} study this problem in the context of investment strategies. 
\cite{SuYu19} study a sequential investment game in a duopoly
setting in the presence of competition and dynamic learning opportunities.

\section{Empirical Bayes Prior} \label{EB}
We now present our model assumptions, and derive an empirical Bayes prior estimator.  

\subsection{Empirical Prior Derivation} \label{EB:Der}
Our problem of interest is one where features can be
grouped into two or more groupings. For example, in a recommendation setting, one can 
distinguish between item and user features. In a personalization setting, one can
distinguish between non-interaction features (e.g., \lq\lq gender\rq\rq) and interaction features 
(e.g., \lq\lq female user likes action movies\rq\rq).

By grouping the features into non-overlapping categories, we impose a Bayesian hierarchical 
model. We assume that each category $C_k$ has a distinct hyperparameter meta-prior distribution
with mean $\nu_k$ and variance $\tau_k^2$. The meta-prior distribution can be determined using experts' knowledge for each specific application. In our experiments, we set the meta-prior distribution to be Gaussian
as explained in section~\ref{EB:Est}. 
We assume that each feature's true effect $\mu_i$ is drawn i.i.d 
(independent and identically-distributed) from that feature's category meta-prior, thus:
\begin{equation} \label{meta-prior-dist}
 \E[\mu_i]=\nu_k, \quad \V[\mu_i]=\tau_k^2, \quad \forall i \in C_k.
\end{equation}
$\E[.]$ and $\V[.]$ refer to the expectation and variance operators 
over random draws $i \in C_k$.

Let $\tilde{\mu}_{i}$ and $\tilde{\sigma}_{i}^2$ respectively denote the estimated (observed) feature
effect and variance for feature $i$. Our approach assumes the existence of a model 
(like a Bayesian regression) that estimates $\tilde{\mu}_{i}$ and $\tilde{\sigma}_{i}^2$.
We assume $\tilde{\mu}_{i}|\mu_{i}$ is drawn from a distribution 
with a mean equal to the true effect $\mu_i$, and variance equal to its estimated variance $\tilde{\sigma}_{i}^2$. 
Based on our assumptions, the following holds for any feature $i \in C_k$:
\begin{equation}\label{derivation-assump}
\E[\tilde{\mu}_{i}|\mu_{i}] = \mu_{i}, \; \V[\tilde{\mu}_{i}|\mu_{i}]=
			                  \tilde{\sigma}_{i}^2, \; \E[\tilde{\mu}_{i}] = \nu_k, \quad \forall i \in C_k.
\end{equation}

For each feature in category $C_k$, we perform variance decomposition:
\begin{align}\label{var_decomp}
 \V[\tilde{\mu}_{i}]  &= \E[\V[\tilde{\mu}_{i}|\mu_{i}]] + \V[\E[\tilde{\mu}_{i}|\mu_{i}]] \nonumber \\
  &= \E[\tilde{\sigma}_{i}^2] + \tau_k^2,  \quad   \forall i \in C_k.
\end{align}
The second equality follows from our assumptions provided in 
Equations~(\ref{meta-prior-dist}) and (\ref{derivation-assump}). Note that 
the choice of meta-prior distribution is irrelevant in deriving 
Equation~(\ref{var_decomp}).

We can  now solve for the meta-prior variance $\tau_k^2$ by:
\begin{equation}\label{priorVar}
 \tau_k^2 = \V[\tilde{\mu}_{i}] - \E[\tilde{\sigma}_{i}^2].
\end{equation} 
We can estimate $\tau_k^2$ for each category
$C_k$ using Equation~(\ref{priorVar}) as long as we have a procedure to estimate 
the feature effect $\tilde{\mu}_{i}$ and its variance
$\tilde{\sigma}_{i}^2$ for any feature $i$ in category $C_k$.
We can obtain estimates of $\E[\tilde{\sigma}_{i}^2]$ and $\V[\tilde{\mu}_{i}]$
by their corresponding sample mean and sample variance, respectively. 
In the next two sections, we provide details of our model and framework for
estimating $\tilde{\mu}_{i}$ and $\tilde{\sigma}_{i}^2$ and propose 
an estimator for the meta-prior variance, $\tau_k^2$. 
%
\subsection{Model Description}\label{sec:model}
%
Our application of interest is a Bayesian generalized linear bandit, which interacts with 
the environment and collects data, while updating its estimates of $\tilde{\mu}_{i}$ and $\tilde{\sigma}^2_{i}$
at each time $t$. In general, these bandits model each feature effect using an underlying Gaussian 
distribution, starting with a standard normal non-informative prior \citep{GLMBandits10,ChLi11}. 
Let $\mathcal{N}(\tilde{\mu}_{i,t}, \tilde{\sigma}_{i,t}^2)$ be the model weight 
distribution associated with feature $i$ at time $t$.
As time progresses, the bandit learns from its interactions with 
the environment, updating the features' weight distributions accordingly. 
In a stochastic setting, as $t \rightarrow \infty$, 
we have $\tilde{\sigma}_{i,t}^2 \rightarrow 0$ and 
$\tilde{\mu}_{i,t} \rightarrow \mu_i$, where $\mu_i$ is the true feature 
effect \citep{BuCe12}.

In our implementation, we use the Bayesian probit regression of \citet{GrCa10},
referred to as the \textbf{B}ayesian \textbf{LI}near \textbf{P}robit (BLIP) model. 
Its probability distribution is given by:
\begin{equation}\label{eq:BLIP}
P(y|x,\tilde{\mu}) = \Phi \left(y \cdot \frac{\tilde{\mu}^Tx}{\beta}\right), 
\end{equation}
where $\Phi(\cdot)$ denotes the CDF of the standard Gaussian distribution and $\beta$ 
scales its steepness. For simplicity, we set $\beta = 1$ and drop it in the remainder of the paper.
Here, $y \in \{-1,1\}$ is the response variable.
Moreover, $x$ denotes the feature vector and $\tilde{\mu}$ refers to its associated weight vector.

We learn the feature weights $\tilde{\mu}$ in a Bayesian fashion. 
The weights are modeled as mutually independent random variables (hence the diagonal coavariance matrix) 
following a Gaussian posterior distributions $\mathcal{N}(\tilde{\mu}, \tilde{\sigma}^2)$. 
The model uses a conjugate standard Gaussian prior $\mathcal{N}(0, 1)$ for the weights.
The weight variances are upper bounded by the initial prior configuration, as every subsequent weight update is guaranteed to not increase the variance \citep{GrCa10}. 

We refer to this GLM classifier as BLIP. BLIP can be used as the Bayesian underlying model for a Thompson Sampling bandit, as we do in Section~\ref{sec:liveExp}. We refer to such a bandit as a BLIP-based TS bandit. Next, we discuss how to apply our empirical Bayes method to BLIP, generating an EB BLIP classifier and an EB BLIP TS bandit.

%
\subsection{Empirical Prior Estimation} \label{EB:Est}
%

For our use-case, we specify the meta-prior distribution as a Gaussian, the conjugate prior of the feature weight distribution. Each feature's true effect $\mu_i$ is drawn from that feature's category meta-prior: 
\begin{equation}
\mu_i \sim \mathcal{N}(\nu_k, \tau_k^2), \quad \forall i \in C_k.
\end{equation}

Let $\hat{\tau}_{k,t}^2$ denotes an estimator for the meta-prior variance
$\tau_{k}^2$ at time $t$. By applying Equation~(\ref{priorVar}), we have
\begin{equation} \label{eq:BayesVarianceWithMean} 
   \hat{\tau}_{k,t}^2 = \widehat{\V[\tilde{\mu}_{i,t}]}-
                            \widehat{\E[\tilde{\sigma}_{i,t}^2]} 
                      = \frac{\sum_{i\in C_k} (\tilde{\mu}_{i,t} - \hat{\nu}_{k,t})^2}{N_k - 1} 
										        -\frac{\sum_{i\in C_k}  {\tilde{\sigma}_{i,t}^2}}{N_k},
\end{equation}
where $N_k$ refers to the number of features in category $C_k$
and superscript $\hat{.}$ notation represents an estimator. 
We estimate $\widehat{\E[\tilde{\sigma}_{i,t}^2]}$ using the empirical mean of the observed variances.
We use the basic sample variance formula for $\widehat{\V[\tilde{\mu}_{i,t}]}$.  $\hat{\nu}_{k,t}$ is the empirical mean estimation of the meta-prior mean $\nu_k$ computed at time $t$ for category $C_k$:
\begin{equation}\label{eq:nu}
 \hat{\nu}_{k,t} = \frac{\sum_{i\in C_k} \tilde{\mu}_{i,t}}{N_k}, \quad \forall C_k. 
\end{equation}
Equation~(\ref{eq:BayesVarianceWithMean}) provides a formula
for estimating the meta-prior variance $\tau^2_k$ at time $t$
per category $C_k$. $\sum_i(\tilde{\mu}_{i,t} - \hat{\nu}_{k,t})^2/(N_k-1)$ 
is the unbiased estimator of the true parameter $\tau^2_k$ with estimation
noise $\sum_i {\tilde{\sigma}_{i,t}^2}/N_k$. 

In our experiments, we set $\nu_k = 0$ to ensure the model is invariant
to input feature sign changes. As long as one includes a bias (intercept) term
in the generalized linear model, setting $\nu_k$ to any value has little effect as its 
value will be absorbed into the bias term. We indeed confirmed this hypothesis
in preliminary experiments, where we compared setting $\nu_k = 0$
to setting it using Equation (\ref{eq:nu}). Setting $\nu_k=0$ gains one degree of freedom,
ensuring the sample variance denominator is $N_k$ and not $N_k-1$. 
Hence, Equation (\ref{eq:BayesVarianceWithMean}) simplifies to:
\begin{equation}\label{eq:BayesVariance}
 \hat{\tau}_{k,t}^2 = 
          \frac{\sum_{i\in C_k}[\tilde{\mu}_{i,t}^2-\tilde{\sigma}_{i,t}^2]}{N_k}, \quad \forall C_k.
\end{equation}
To ensure a non-degenerative $\tau_k^2$, one can enforce a minimum value threshold (see Section\ref{sec:Bootstrapping} for additional approaches).
Our method is a parametric (zero-mean Gaussian) g-modeling EB approach \citep{EfHa16}
where we aim at estimating the variance $\tau_k^2$. 

In practice, we start the model with a non-informative prior.
At some small $t$ (even at the end of 
batch $t=1$, where all the data is random), we
compute the empirical Bayes prior $\mathcal{N}(0,\hat{\tau}_{k,t}^2)$ 
for each feature grouping $k$ using 
Equation~(\ref{eq:BayesVariance}). 
We then restart the model using the new informative prior
and re-train it using the data from the elapsed $t$ 
batches. One can repeat these steps at multiple time-points $t$, each time re-computing a new $\tau_k^2$
in an expectation-maximization fashion,
but the approximation after one round is likely sufficient.

The computational complexity overhead of this method is negligible, 
as the meta-prior is computed and applied only once.
Computing $\hat{\tau}_k^2$ in Equation~(\ref{eq:BayesVariance}) is $O(N)$, 
linear in the total number of features $N$. 
Once $\hat{\tau}_k^2$ is computed, one needs to re-train with data up to time $t$. The time complexity
of such batch update is model dependent, and can be as small as $O(N)$ if a closed-form formula exists for 
already-processed batch updates. If one has to actually re-process the data, 
the training time complexity would be 
$O(tN)$, where $t$ is a small constant $t << T$. In the latter case, the batch re-training can take place offline to further reduce the time overhead. In fact, most production systems have regular batch updates at pre-determined intervals, and this update can simply tag along the regular schedule.

Note that our derivation of $\hat{\tau}_{k,t}^2$ in this section is general 
and applicable to any Bayesian learning framework, and  
not tied to our specific bandit user-case. One can estimate the meta-prior
variance using Equation~(\ref{eq:BayesVarianceWithMean}) as long as 
estimates of the feature effect and its variance are available. 

\section{Theoretical Results}\label{sec:theory} 
In this section, we first provide statistical properties of 
our meta-prior variance estimator. We then report an upper bound
on the regret of our TS-based GLM bandit model.   
%
\subsection{Empirical Prior: Unbiasedness Property} \label{EB:prop}
%
In this section, we prove that the meta-prior variance estimator,
$\hat{\tau}_{k}^2$, given in Equation~(\ref{eq:BayesVarianceWithMean}),
is unbiased as long as the estimates of $\tilde{\mu}_{i}$ are uncorrelated for all $i \in C_k$.
The meta-prior variance can be computed at any time $t$ and the results in this 
section hold for any time selected. Therefore, for clarity of notation,
we drop the subscript $t$ in our analysis.  

\begin{proposition}\label{thm:proposition1}
Let the meta-prior variance estimator, $\hat{\tau}_{k}^2$, for each category $C_k$, be  
\begin{equation} \label{eq:proposition1} 
  \hat{\tau}_{k}^2 = \widehat{\V[\tilde{\mu}_{i}]}-
                            \widehat{\E[\tilde{\sigma}_{i}^2]} 
                      = \frac{\sum_{i\in C_k} (\tilde{\mu}_{i} - \hat{\nu}_{k})^2}{N_k - 1} 
										        -\frac{\sum_{i\in C_k}  {\tilde{\sigma}_{i}^2}}{N_k}, 
\end{equation}
where $\tilde{\mu}_{i}$ and $\tilde{\sigma}_{i}^2$ respectively denote the estimated feature
effect, and estimated variance, for feature $i$. The $\E[.]$ and $\V[.]$ operators  
refer to the expectation and variance over random draws $i \in C_k$. Moreover, 
$\hat{\nu}_{k}=\frac{\sum_{i\in C_k} \tilde{\mu}_{i}}{N_k}$ represents the empirical
mean estimation of the meta-prior mean $\nu_k$, and $N_k$ refers to the number of 
features in category $C_k$. We have: 
\begin{equation} \label{prior-bias}
 \E[\hat{\tau}_{k}^2] = \tau_{k}^2-\frac{1}{N_k(N_k-1)}\sum_{\substack{i,j\in C_k \\ i \neq j}}
											    \Cov[\tilde{\mu}_{i},\tilde{\mu}_{j}].
\end{equation}
\end{proposition}

\proof{Proof}
One can re-write $\widehat{\V[\tilde{\mu}_{i}]}$ as: 
\begin{equation}
 \widehat{\V[\tilde{\mu}_{i}]} = \frac{1}{N_k - 1}\sum_{i\in C_k} (\tilde{\mu}_{i} - \hat{\nu}_{k})^2
                    = \frac{1}{N_k - 1} \Big[ \sum_{i\in C_k} \tilde{\mu}_{i}^2-N_k\hat{\nu}_{k}^2 \Big].
\end{equation}
Taking the expectation on both sides and applying the linearity of expectation gives: 
\begin{align}\label{varBias}
 \E[\widehat{\V[\tilde{\mu}_{i}]}] 
     &=\frac{1}{N_k-1} \Big[ \sum_{i\in C_k} \E[\tilde{\mu}_{i}^2]-N_k \E[\hat{\nu}_{k}^2] \Big] \nonumber \\
	   &= \frac{1}{N_k-1} \bigg[ \sum_{i\in C_k} \Big( \E[\tilde{\mu}_{i}]^2+\V[\tilde{\mu}_{i}] \Big)
		     -N_k \Big( \E[\hat{\nu}_{k}]^2+\V[\hat{\nu}_{k}] \Big) \bigg] \nonumber \\
	   &= \frac{1}{N_k-1} \bigg[ \sum_{i\in C_k} \Big( \nu_k^2+\V[\tilde{\mu}_{i}] \Big)
				-N_k \Big( \nu_k^2 + \frac{1}{N_k^2} \Big( \sum_{i\in C_k} \V[\tilde{\mu}_{i}]
	         +\sum_{\substack{i,j\in C_k \\ i \neq j}} \Cov[\tilde{\mu}_{i}, \tilde{\mu}_{j}] 
					   \Big) \Big) \bigg] \nonumber \\
		&= \frac{1}{N_k}\sum_{i\in C_k} \V[\tilde{\mu}_{i}] - 
          \frac{1}{N_k(N_k-1)}\sum_{\substack{i,j\in C_k \\ i \neq j}}\Cov[\tilde{\mu}_{i},\tilde{\mu}_{j}]
					        \nonumber \\
		&= \frac{1}{N_k}\sum_{i\in C_k} (\tau_k^2+\E[\tilde{\sigma}_{i}^2]) - 
          \frac{1}{N_k(N_k-1)}\sum_{\substack{i,j\in C_k \\ i \neq j}}\Cov[\tilde{\mu}_{i},\tilde{\mu}_{j}]
					        \nonumber \\
		&= \tau_{k}^2 + \E \Bigg[ \frac{\sum_{i\in C_k}\tilde{\sigma}_{i}^2}{N_k} \Bigg]
		    - \frac{1}{N_k(N_k-1)}\sum_{\substack{i,j\in C_k \\ i \neq j}}\Cov[\tilde{\mu}_{i},\tilde{\mu}_{j}].
\end{align}
The second equality follows from applying the variance formula ($\V[\mu] = \E[\mu^2] - \E[\mu]^2$) 
to the random variables $\tilde{\mu}_{i}$ and $\hat{\nu}_{k}$. 
The third equality follows from one of our assumptions provided in Equation (\ref{derivation-assump}), 
$\E[\tilde{\mu}_{i}]=\nu_{k}\ \forall i \in C_k$, and the definition of 
$\hat{\nu}_{k}=\frac{\sum_{i\in C_k} \tilde{\mu}_{i}}{N_k}$. 
In the penultimate equality, we substitute the value of 
$\V[\tilde{\mu}_{i}]$ using Equation~(\ref{priorVar}).

Taking the expectation of $\hat{\tau}_{k}^2$ in~Equation (\ref{eq:proposition1}) and substituting 
$\E[\widehat{\V[\tilde{\mu}_{i}]}]$ from Equation~(\ref{varBias}) gives: 
\begin{equation} \label{priorBias}
 \E[\hat{\tau}_{k}^2] = \E \Bigg[ \frac{\sum_{i\in C_k} (\tilde{\mu}_{i}-\hat{\nu}_{k})^2}{N_k-1} \Bigg] 
										         -\E \Bigg[ \frac{\sum_{i\in C_k} {\tilde{\sigma}_{i}^2}}{N_k} \Bigg] 
						          = \tau_{k}^2-\frac{1}{N_k(N_k-1)}\sum_{\substack{i,j\in C_k \\ i \neq j}}
											    \Cov[\tilde{\mu}_{i},\tilde{\mu}_{j}], 
\end{equation}
which concludes our proof. \Halmos
\endproof

\begin{corollary}\label{thm:corollary1}
 The meta-prior variance estimator $\hat{\tau}_{k}^2$ for each category $C_k$,
 given in Equation (\ref{eq:proposition1}), is unbiased when the estimates
 of $\tilde{\mu}_{i}$ and $\tilde{\mu}_{j}$, for all $i,j\in C_k$ where $i \neq j$,
 are uncorrelated. 
\end{corollary}

\proof{Proof}
This corollary simply follows from the results of Proposition \ref{thm:proposition1}.
Since $\tilde{\mu}_{i}$ and $\tilde{\mu}_{j}$ are uncorrelated, 
$\Cov[\tilde{\mu}_{i},\tilde{\mu}_{j}]=0$ for all $i,j\in C_k$
where $i \neq j$. Applying Equation (\ref{prior-bias}) gives $\E[\hat{\tau}_{k}^2] = \tau_{k}^2$
which proves the unbiasedness of the meta-prior variance estimator $\hat{\tau}_{k}^2$. \Halmos
\endproof

\begin{corollary}\label{thm:corollary2}
 The meta-prior variance estimator $\hat{\tau}_{k}^2$ for each category $C_k$,
 given in Equation~(\ref{eq:proposition1}), is unbiased when applied to our Bayesian GLM 
 bandit model presented in Equation~(\ref{eq:BLIP}).
\end{corollary}

\proof{Proof}
This result holds due to the modeling assumptions of 
Section~\ref{sec:model}. We assume that feature weights
are mutually independent which results in a diagonal covariance matrix, 
making the estimates of $\tilde{\mu}_{i}$ uncorrelated. Therefore by 
Proposition~\ref{thm:proposition1} and Corollary~\ref{thm:corollary1}, 
our meta-prior variance estimator in our Bayesian GLM model is unbiased. \Halmos
\endproof

\subsection{Empirical Prior: Strong Consistency}
In this section, we show strong consistency of our proposed meta-prior 
variance estimator in the regime where the estimates of $\tilde{\mu}_{i}$ and $\tilde{\sigma}_{i}^2$
are independent for all $i \in C_k$, and random variables $\tilde{\mu}_{i}$,
$\tilde{\mu}_{i}^2$, and $\tilde{\sigma}_{i}^2$ have finite expectations and variances. 

\begin{proposition} 
The meta-prior variance estimator $\hat{\tau}_{k}^2$,
given in Equation~(\ref{eq:proposition1}), is strongly consistent
when the feature effect estimates $\tilde{\mu}_{i}$ and their observed variances 
$\tilde{\sigma}_{i}^2$ are independent $\forall i\in C_k$,
and the expectation and variance exist for random variables $\tilde{\mu}_{i}$,
$\tilde{\mu}_{i}^2$, and $\tilde{\sigma}_{i}^2$.
\end{proposition}

\proof{Proof} 
The strong consistency of our meta-prior variance 
estimator follows from the Kolmogorov Strong Law of Large Numbers; refer to 
Theorem 2.3.10 in~\cite{SeSi94} for details and proof. 

\begin{theorem} [Kolmogorov Strong Law of Large Numbers] (Thm 2.3.10 in \cite{SeSi94})
Let $X_i$, $i \ge 1$, be independent random variables such that $\E[X_i]=\mu_i$
and $\V[X_i]=\sigma_i^2$ exists for every $i \ge 1$. Also let, 
$\bar{X}_n=\frac{\sum_{i=1}^{n}{X_i}}{n}$ and 
$\bar{\mu}_n=\frac{\sum_{i=1}^{n}\mu_i}{n}$ for $n \ge 1$. Then,
\[
 \sum_{k\ge 1} k^{-2} \sigma_k^2<\infty \quad \Longrightarrow \quad \bar{X}_n \xrightarrow{a.s.} \bar{\mu}_n.
\]
\end{theorem}

To prove strong consistency of meta-prior variance estimator
$\hat{\tau}_{k}^2$, we first rewrite the sample variance estimator 
$\widehat{\V[\tilde{\mu}_{i}]}$ and apply the definition of $\hat{\nu}_{k}$:
\begin{equation} 
\widehat{\V[\tilde{\mu}_{i}]} 
           = \frac{\sum_{i\in C_k} (\tilde{\mu}_{i} - \hat{\nu}_{k})^2}{N_k-1} 
					 = \frac{\sum_{i\in C_k} \tilde{\mu}_{i}^2}{N_k-1}-\frac{N_k}{N_k-1}\hat{\nu}_{k}^2 \nonumber \\
				   = \frac{N_k}{N_k-1} \Bigg[ \frac{\sum_{i\in C_k} \tilde{\mu}_{i}^2}{N_k}
								-\Big(\frac{1}{N_k}\sum_{i\in C_k} \tilde{\mu}_{i} \Big)^2 \Bigg ].
\end{equation}
Hence the meta-prior variance estimator $\hat{\tau}_{k}^2$,
given in Equation~(\ref{eq:proposition1}), becomes: 
\begin{equation} \label{eq:proposition2} 
 \hat{\tau}_{k}^2 = \frac{N_k}{N_k-1} \Bigg[ \frac{\sum_{i\in C_k} \tilde{\mu}_{i}^2}{N_k}
								-\Big(\frac{1}{N_k}\sum_{i\in C_k} \tilde{\mu}_{i} \Big)^2 \Bigg ]
								  -\frac{\sum_{i\in C_k}  {\tilde{\sigma}_{i}^2}}{N_k}.
\end{equation}
In the following, we provide details on how to apply Kolmogorov Strong Law of Large Numbers
on the summation terms with random variables $\tilde{\mu}_{i}$, $\tilde{\mu}_{i}^2$, and 
$\tilde{\sigma}_{i}^2$ and prove strong consistency of $\hat{\tau}_{k}^2$.

Given our assumptions, estimates $\tilde{\mu}_{i} \ \forall i\in C_k$ are independent 
and the expectation of $\tilde{\sigma}_{i}^2$ exists. The latter implies
$\exists \epsilon  \textnormal{ s.t. }  \E[\tilde{\sigma}_{i}^2] \le \epsilon,  \forall i\in C_k$.
We further 
have $\E[\tilde{\mu}_{i}] = \nu_k$ per our assumption in Equation~(\ref{derivation-assump}), 
and Equation~(\ref{var_decomp}) results in $\V[\tilde{\mu}_{i}] = \tau_k^2+\E[\tilde{\sigma}_{i}^2] \le \tau_k^2 + \epsilon$. Note that $\sum_{i \ge 1}^\infty \frac{1}{i^2} = \frac{\pi^2}{6}$. We can now apply the
Kolmogorov strong law of large numbers on $\tilde{\mu}_{i}$:
\begin{equation}\label{SLLN1}
\sum_{i\ge 1} i^{-2}\V[\tilde{\mu}_{i}] = 
	      \sum_{i\ge 1} \frac{\tau_{k}^2+\E[\tilde{\sigma}_{i}^2]}{i^2} 
       	\le \frac{\pi^2}{6}(\tau_k^2+\epsilon) < \infty
       	\quad \Longrightarrow \quad
 \frac{1}{N_k}\sum_{i\in C_k} \tilde{\mu}_{i} \ \ \xrightarrow{a.s.} \ \ 
               \frac{1}{N_k}\sum_{i\in C_k}\E[\tilde{\mu}_{i}] = \nu_k.       	
\end{equation}

We apply similar arguments to the independent random variables 
$\tilde{\mu}_{i}^2,  \forall i\in C_k$. We have 
\[
 \E [\tilde{\mu}_{i}^2] = \V [\tilde{\mu}_{i}]+\E [\tilde{\mu}_{i}]^2
                          = \tau_{k}^2+\E[\tilde{\sigma}_{i}^2]+\nu_k^2, 
\]
using the variance formula, our assumption in Equation~(\ref{derivation-assump}), and Equation~(\ref{var_decomp}). 
Moreover, we assume that $\V[\tilde{\mu}_{i}^2]$ exists which implies that 
$\exists \epsilon^\prime  \textnormal{ s.t. } \V[\tilde{\mu}_{i}^2] \le \epsilon^\prime,  \forall i\in C_k$.
Therefore, $\sum_{i\ge 1} i^{-2}\V[\tilde{\mu}_{i}^2]=O(1)$. 
Hence, by the Kolmogorov strong law of large numbers, as $N_k \rightarrow \infty$, almost surely 
\begin{equation}\label{SLLN2}
 \frac{1}{N_k}\sum_{i\in C_k} \tilde{\mu}_{i}^2 \ \ \xrightarrow{a.s.} \ \ 
     \frac{1}{N_k}\sum_{i\in C_k} \E[\tilde{\mu}_{i}^2]=\tau_{k}^2+\nu_k^2
		             + \frac{1}{N_k} \sum_{i} \E[\tilde{\sigma}_{i}^2].
\end{equation}
By applying the results in Equations (\ref{SLLN1}) and (\ref{SLLN2}), we obtain 
\begin{align}\label{SLLN:varEst}
 \lim_{N_k \to \infty} \widehat{\V[\tilde{\mu}_{i}]} 
       &= \lim_{N_k \to \infty} \frac{N_k}{N_k-1}
			   \Big[ \lim_{N_k \to \infty} \frac{\sum_{i\in C_k} \tilde{\mu}_{i}^2}{N_k}
					-\Big( \lim_{N_k \to \infty} \frac{1}{N_k}\sum_{i\in C_k} \tilde{\mu}_{i} \Big)^2 \Big] \nonumber \\
			 &= \tau_{k}^2+\nu_k^2+\frac{1}{N_k} \sum_{i} \E[\tilde{\sigma}_{i}^2]-\nu_k^2  \nonumber \\
			 &= \tau_{k}^2+\frac{1}{N_k} \sum_{i} \E[\tilde{\sigma}_{i}^2].
\end{align}
Finally, by taking the limit of the meta-prior variance estimator $\hat{\tau}_{k}^2$,
given in Equation (\ref{eq:proposition1}), we obtain 
\begin{align*}
   \lim_{N_k \to \infty} \hat{\tau}_{k}^2 &= \lim_{N_k \to \infty} 
	            \widehat{\V[\tilde{\mu}_{i}]}-\lim_{N_k \to \infty} \widehat{\E[\tilde{\sigma}_{i}^2]} \\
							&= \tau_{k}^2+\frac{1}{N_k} \sum_{i} \E[\tilde{\sigma}_{i}^2] 
							             - \lim_{N_k \to \infty} \frac{\sum_{i\in C_k}{\tilde{\sigma}_{i}^2}}{N_k} \\
							&= \tau_{k}^2+\frac{1}{N_k} \sum_{i} \E[\tilde{\sigma}_{i}^2] 
							             - \frac{1}{N_k} \sum_{i} \E[\tilde{\sigma}_{i}^2] \\
							&= \tau_{k}^2, 
\end{align*}
where the second equality applies the results in (\ref{SLLN:varEst}). 
The third equality follows from the Kolmogorov strong law of large numbers applied 
to the independent random variables $\tilde{\sigma}_{i}^2 \ \forall i\in C_k$ that 
have finite expectation and variance due to our assumptions. This concludes the strong 
consistency proof of our proposed meta-prior variance estimator. \Halmos
\endproof
%
\subsection{Bandit Cumulative Regret Upper Bound}\label{sec:regret}
%
In this section, we provide an upper bound on the regret 
of our EB BLIP-based bandit method, described in Section~\ref{sec:model}. 
In the stochastic GLM bandit setting, at every round $t$, we sample $\dot{\mu} \sim \tilde{\mu}_t$, and play arm:
\begin{equation}\label{eq:TS}
x_t(\dot{\mu}) = \argmax_{x \in X}\ \Phi \left(\dot{\mu}^T x \right).
\end{equation}
Let us denote the optimal weight vector as $\mu^*$ and the optimal arm as $x^* = x(\mu^*)$.
The instantaneous regret at time step $t$ is the difference between the expected reward 
of the optimal arm $x^{*}$ and the selected arm $x_t$, written as 
$\Phi \left({\mu^{*}}^T x^{*} \right) - \Phi \left({\mu^{*}}^T x_t \right)$.
Over a time horizon of length $T$, the cumulative regret becomes:
\begin{equation}\label{eq:cumRegret}
 R^{GLM}(T) = \sum_{t=1}^{T} \Big[ \Phi \left({\mu^{*}}^T x^{*} \right) 
                                  - \Phi \left({\mu^{*}}^T x_t \right) \Big].
\end{equation}

Following Lemma 4 of \cite{AbLa17}, and under a set of assumptions, the cumulative regret $R^{GLM}(T)$ is upper bounded. We restate the Lemma for completeness and report on our regret bounds 
in the two corollaries that follow afterward. 

\begin{lemma} \label{Lemma:regretBound}
(Lemma 4 of \citet{AbLa17})\\
Assume that:
\begin{enumerate}
\item The arm set $X$ is a bounded closed subset of $\mathbb{R}^d$
such that $\| x \|\leq 1, \ \forall x \in X$.
\item There exists a known $S \in \mathbb{R}^{+}$ such that $\|\mu^* \| \le S$.
\item The noise process is a martingale difference sequence given a filtration, and is conditionally $R$-subgaussian for some constant $R \geq 0$.
\item The link function $\Phi$ is continuously differentiable, Lipschitz with constant $k_\Phi$, and its derivative is lower bounded by $c_\Phi =\inf_{\mu \in \mathbb{R}^d, x \in X} \Phi'(\mu^T x) >0$.
\end{enumerate}
Then, with probability $1-\delta$, the cumulative regret $R^{GLM}(T)$ of the GLM TS bandit is upper bounded by:
\begin{equation}\label{eq:regretGLM}
 R^{GLM}(T) \le \frac{k_\Phi}{c_\Phi}(\beta_T(\delta')
             + \gamma_T(\delta')(1+2/p))\sqrt{2Td\log\bigg(1+\frac{T}{\lambda}\bigg)}
						 + \frac{2k_\Phi\gamma_T(\delta')}{pc_\Phi}\sqrt{\frac{8T}{\lambda}\log\frac{4}{\delta}}.
\end{equation}
Here $\delta'=\frac{\delta}{4T}$, $\lambda$ denotes the 
regularization parameter, $p$ is an anti-concentration probability bound,
\begin{equation}\label{eq:regretConst}
 \beta_t(\delta) = R\sqrt{2\log\frac{(\lambda+t)^{d/2}\lambda^{-d/2}}{\delta}}+\sqrt{\lambda}S, 
 \quad \gamma_t(\delta) = \beta_t(\delta') \sqrt{cd\log(c'd / \delta)},
\end{equation}
and $c, c'$ are positive concentration constants.
\end{lemma}

\begin{corollary}\label{thm:corollary3}
The cumulative regret of the EB BLIP-based TS bandit
is upper bounded by Equation~(\ref{eq:regretGLM}) and is thus of order of 
$\tilde{O}(d^{3/2} \sqrt{T})$.
\end{corollary}

\proof{Proof}
Our approach ascribes to the Lemma~\ref{Lemma:regretBound} assumptions, and is hence bounded. 
Assumption 1 is satisfied by using a 1-hot encoding for categorical features, and centering and scaling numerical features. In practice, one can take a large $S$ to satisfy Assumption 2. As our reward is binary, Assumption 3 is satisfied, with $R = 1/2$. For Assumption 4, our $\Phi(.)$ is the CDF of the standard Gaussian distribution:
\begin{equation}
 \Phi(x)=\frac{1}{2} \bigg[ 1+erf\bigg(\frac{x}{\sqrt{2}}\bigg) \bigg], 
   \quad \textrm{where} \quad erf(x)=\frac{2}{\sqrt{\pi}} \int_{0}^{x} e^{-t^2}dt.
\end{equation}
By taking the derivative of $\Phi(x)$ w.r.t $x$ and substituting for 
$\frac{d}{dx} erf(x)= \frac{2}{\sqrt{\pi}} e^{-x^2}$, we see that it is 
continuously differentiable:
$
 \frac{d\Phi(x)}{dx}=\frac{1}{\sqrt{2\pi}} e^{-\frac{x^2}{2}}.
$
By the mean-value theorem, any differentiable function with
bounded derivative is Lipschitz. $\Phi(.)$ is Lipschitz,
since its first derivative is bounded by 
$0 < \frac{d\Phi(x)}{dx} < \frac{1}{\sqrt{2\pi}}$, giving us a suitable value for 
$k_\Phi = \frac{1}{\sqrt{2\pi}}$. Note that $c_\Phi \rightarrow 0^+$ as $\mu^T x \rightarrow \pm \infty$. 
As both $x$ and $\mu$ are bounded by assumptions 1 and 2, this guarantees $c_\Phi > 0$. 

Now that all four assumptions are satisfied, Lemma~\ref{Lemma:regretBound} holds, 
and the cumulative regret of our EB BLIP bandit is bounded according to Equation~(\ref{eq:regretGLM}). 
This bound is of order of $\tilde{O}(d^{3/2} \sqrt{T})$. \halmos

\endproof

\begin{corollary}\label{thm:corollary4}
The cumulative regret of the BLIP-based TS bandit is of order
$\tilde{O}(d^{3/2} \sqrt{T})$, whether an empirical Bayes
prior $\mathcal{N}(0, \tau_k^2)$ per feature category $C_k$ 
is applied, or a non-informative prior $\mathcal{N}(0,1)$ is used. 
\end{corollary}

\proof{Proof}
Following the results in Lemma~\ref{Lemma:regretBound} and 
Corollary~\ref{thm:corollary3}, we obtain a frequentist regret 
bound of order $\tilde{O}(d^{3/2} \sqrt{T})$. This bound is optimal 
for GLM TS bandits \citep{hamidi2020worstcase}. 
This bound applies to our bandit, regardless if we use the empirical Bayes prior 
$\mathcal{N}(0, \tau_k^2)$, 
or the non-informative prior $\mathcal{N}(0,1)$. The difference between these 
two alternatives resides in different values of the anti-concentration ($p$) 
and concentration ($c, c'$) constants introduced in \cite{AbLa17}, with no significant
effect on the bound.  

Starting with a $\mathcal{N}(0,1)$ prior, we get anti-concentration probability
$p=\frac{e^{-\frac{1}{2}}}{4\sqrt{\pi}}$ and concentration constants $c=c'=2$, as 
per Appendix A in \cite{AbLa17}. 
We repeat this exercise starting from a $\mathcal{N}(0, \tau_k^2)$ prior 
per feature category $C_k$. 
%
%
Following the same strategy as in the standard Gaussian prior case, and applying 
the Gaussian tail bound provided in \cite{ChCo11}, we get  anti-concentration probability 
$p=\frac{e^{\big(\frac{1}{2}-\frac{1}{\tau_{\min}^2}\big)}}{4\sqrt{\pi}}$ 
and concentration constants $c=2\tau_{\max}^2$ and $c'=2$. Here 
$\tau_{\min}=\min_{k \in C_k} \tau_k$, and $\tau_{\max}=\max_{k \in C_k} \tau_k$. 
For the sake of completeness, we provide 
details on how to compute these values in the appendix. \halmos
\endproof

Let $regretBound$ refer to the r.h.s of Equation~(\ref{eq:regretGLM}). We note that $regretBound \propto \frac{\gamma}{p} \propto \frac{\sqrt{c}}{p} \propto \tau_{\max} \, e^{\frac{1}{\tau^2_{\min}}}$. 
As $\tau_{\min} \rightarrow 0$, $regretBound \rightarrow \infty$. This finding is concordant with previous findings
\citep{HoTa14, LiLi16}. As $\tau_k$ shrinks, exploration is severely hampered, the bandit fails to identify the best arm, and the the cumulative regret grows linearly. As our approach decouples the learning rates of different categories of features, $\tau_k \approx 0$ reflects a category with no variance, and one can use this information to prune or cluster such features. We discuss guardrails in Section~\ref{Discussion}.


The regret bound and asymptotic optimal regret analysis we provide is very specific to the finite horizon bandit problem (i.e., a Markov Decision Process (MDP) with a single state). In general, it is unclear that these methods would generalize to arbitrary Bayesian decision problems. In particular
cases such as tabular MDPs, finite-time regret bounds have been worked out in a frequentist sense 
\citep{SiJa19}.

As for an analysis of exact Bayesian optimality and regret, we refer the reader to Chapter 35 of 
\cite{LaSz20}. As a quick summary, in infinite horizon discounted Bayesian k-armed bandits, Bayesian optimal policy takes the form of a Gittins index when the objective is to maximize total expected discounted reward (Theorem 35.9). For finite horizon bandits, the Bayesian optimal policy is often not an index policy. Computing Bayesian optimal policy for large finite horizon settings is intractable. For small horizon and number of arms, the dynamic programming formulation for achieving the Bayesian optimal policy has a large number of states. But the benefits of obtaining exact optimality may be worth the extensive computation effort in these cases.

\section{Supervised Learning Dataset and Experiments} \label{sec:simulationExp}
In order to validate our method and test its generalization, we first report simulations
on a public optimization dataset from a different domain. 

\subsection{Supervised Learning Dataset}\label{dataPrep}
We pick the Adult dataset 
available from the UCI Machine Learning Repository \citep{BlMe98}.
The target is to predict whether one's income exceeds \$50,000 
per year based on census data. The train and test datasets have 30,162 and 15,060 
observations respectively, after removing rows with empty feature values. 
The dataset has 13 categorical features (see
Table~\ref{dataSet_stats}).


\begin{table}
\centering
 \begin{tabular}{l c | l c} 
 \toprule
   Feature &   \#Values  & Feature & \#Values \\ 
 \midrule
 occupation      &   14   &   workclass        &   7   \\ 
 education       &   16   &   education-num    &   10  \\ 
 relationship    &   6    &   marital status   &   7   \\ 
 gender	         &   2    &   race             &   5   \\ 
 hours-per-week  &   12   &   native-country   &   41  \\ 
 capital gain	 &   3    &   capital loss     &   2   \\ 
 age		&   20   &                &     \\
\bottomrule
 \end{tabular}
\vspace{0.1in}
\caption{Adult dataset features and number of possible values.}
\label{dataSet_stats}
\end{table}


The original 13 features constitute our first-order features. We mimic
our use-case of interest by generating second-order 
features through pairwise combinations of first-order features. 
We add all these second-order 
interaction terms (total of 78) to the Adult dataset, forming our raw feature vector. 
The first-order and second-order groupings 
form a natural categorization for the $\hat{\tau}^2_k$ computation given in Equation~(\ref{eq:BayesVariance}).
%
\subsubsection{Bootstrapping for Small Traffic}\label{sec:Bootstrapping}
To simulate batch updates, we divide the training set equally into 6 batches of 5027 examples
each. We train the probit linear model on the first batch, starting with a $\mathcal{N}(0,1)$ 
prior. We then apply our EB approach described in Section~\ref{EB}
to compute empirical priors $\hat{\tau}^2_1$ over the first-order features,
and $\hat{\tau}^2_2$ over the second-order features. To our initial surprise, this resulted in
degenerative negative $\hat{\tau}^2_k$ values.

This degenerative result stems from an inappropriate initial prior, to our point.
As we had only observed a small number of data points, the default $\mathcal{N}(0,1)$ initial prior
still dominates the first-batch posterior used to compute the empirical Bayes prior. 
The value of $\tilde{\mu}_{i,1}$
was still close to $0$ and the value of $\sigma_{i,1}^2$ close to $1$. 
As $\tilde{\mu}_{i,1} < \tilde{\sigma}_{i,1}$, Equation (\ref{eq:BayesVariance}) 
returns $\hat{\tau}^2_k < 0$. This observation suggests that our first batch is too small 
for empirical prior estimation. 

The necessary traffic required to ensure a non-degenerative $\hat{\tau}^2_k$ depends on the update mechanism of the inference model used, the initial prior, and the magnitude of the true meta-prior variance (see Equation~\ref{eq:BayesVarianceWithMean}). To pursue a model-free discussion, we define large traffic as one where the first update batch is enough to generate viable $\hat{\tau}^2_k > 0, \forall k$. 
Medium traffic is one where $\hat{\tau}^2_k > 0$ within a user-specified number of batches (in our case, one week if one training batch used per day), and small traffic requires waiting even longer.

One solution is to simply consume additional
batches before computing the empirical prior. This ensures that 
$\tilde{\mu}_{i}$ and $\tilde{\sigma}_{i}^2$ deviate enough from their initial values
to produce a viable $\hat{\tau}^2_k$. This solution works well in medium traffic cases, where waiting 
for the next batch can resolve the issue. In small traffic situations, one may not be able to 
wait long enough. 

Here lies the essence of our challenge. Starting from an 
informative prior helps most in small traffic cases and small traffic hinders the proper 
computation of the empirical prior. 
Observe though that the value of  
$\hat{\tau}_k^2$ increases with the number of samples due to the reduction 
of the posterior variance $\tilde{\sigma}^2_{i,1}$. 
One can use techniques like bootstrapping, ensembles, or epoch training to artificially boost the number of samples enough to produce a proper $\hat{\tau}_k^2$. 

For our simulation dataset, and given the first batch of 5K instances, we bootstrap it (sampling with replacement) repeatedly creating multiple sets of 5K instances each. We treat each bootstrapped set as a training epoch.
After each training epoch, we compute $\hat{\tau}_k^2$. We stop training as soon as we attain positive $\hat{\tau}_k^2$ values, which was 8 epochs in our case. We also tried training epochs without bootstrapping, with similar results. 
We keep $\mathcal{N}(0,1)$ as the initial prior, since this is the default non-informative
prior used in many applications. Once we compute the empirical prior $\hat{\tau}^2_k$, 
we restart the model using the new informative prior
and re-train it with the initial non-bootstrapped first batch.

%
\subsubsection{Data Pre-Processing}\label{sec:dataPreprocessing}
One may suspect that not all features in a model are relevant. To that effect,
and to balance the overfitting aspect of small data bootstrapping, 
we prune the bootstrapped model by applying adaptive lasso. 
We pick adaptive lasso due to 
its oracle properties, as it can identify the right subset of features to retain
as if the true underlying model was given in advance \citep{ZouHui06}.

\begin{figure}[!ht]
  \centering 
  \includegraphics[width=0.6\columnwidth]{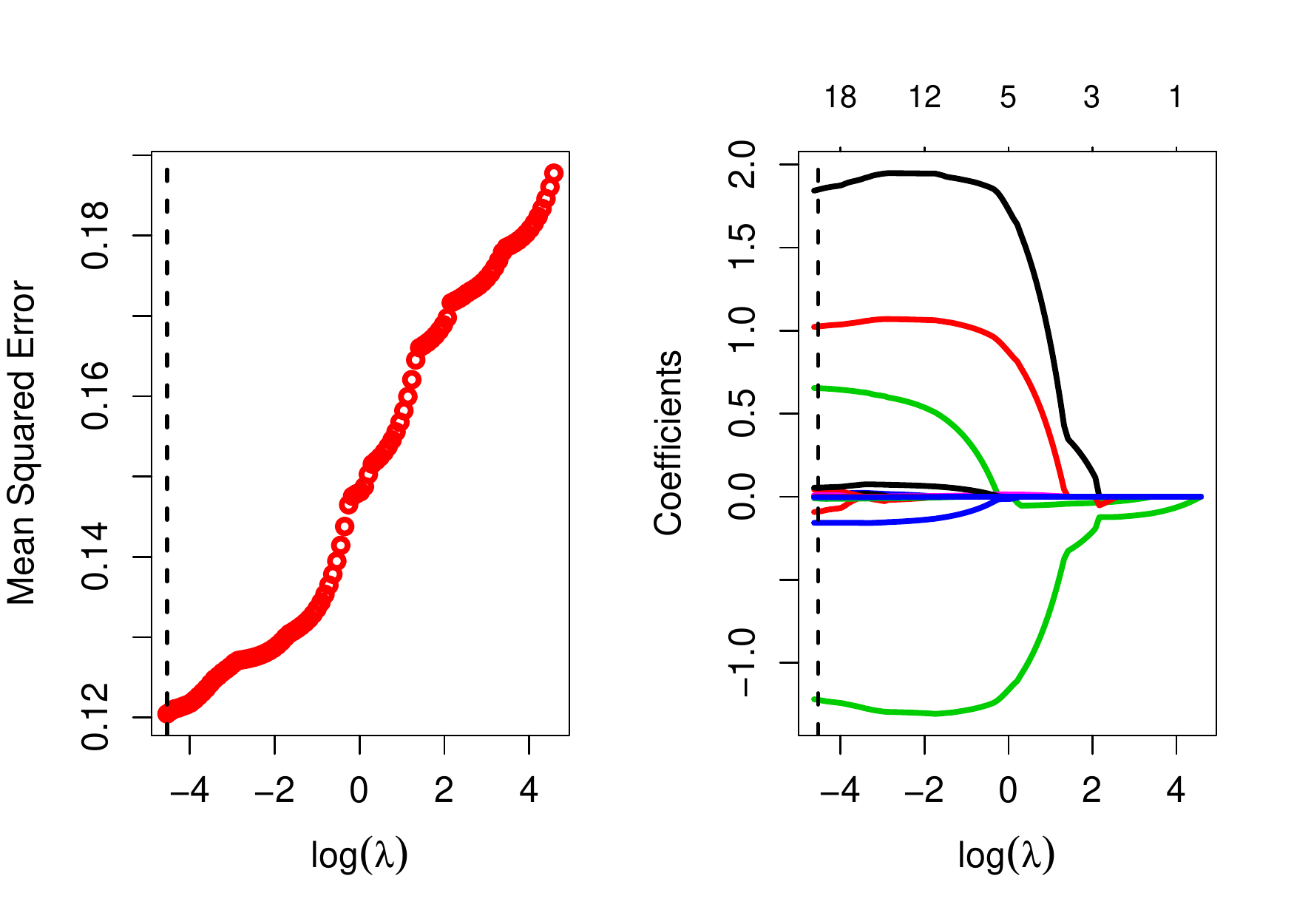}
  \caption{Mean squared error and coefficients selected by adaptive lasso. 
    The left plot shows the mean squared error where the dashed line
    denotes the optimal value of the shrinkage parameter $\lambda$ obtained through
    cross validation.
    The right plot shows the adaptive lasso's pruning order where 
    the top values denote the number of non-zero weight coefficients
    as the shrinkage parameter increases. 
  }
  \label{bootstrap_40k_adlasso_All1st_MSE_Coefs}
\end{figure}

The objective function for adaptive lasso is
\begin{equation}
  \widehat{W} := \argmin_{W}\  ||y-XW||^2 + \lambda \sum_{i} \zeta_{i} |w_{i}|,  
\end{equation}
where $\lambda$ is the shrinkage parameter estimated using
cross validation, $W$ is the model weight vector where each 
$w_{i}$ is the weight coefficient of feature $i$,
$y$ is the response vector, and $X$ the design matrix.
$\zeta$ is the adaptive weight vector and is defined as 
$\zeta_{i} = \frac{1}{|\hat{w}_{i}|^{\gamma}}$, where 
$\hat{w}_{i}$ is an initial estimate of coefficient $w_{i}$
obtained by performing ridge regression in our simulations. 
Moreover, $\gamma$ is a positive constant that adjusts the adaptive lasso
weight vector and is set to $\gamma=1$ in our experiments. 
Figure~\ref{bootstrap_40k_adlasso_All1st_MSE_Coefs} shows an example
of adaptive lasso feature selection using the glmnet package \citep{FrHa10}.

In this section, we use the terms ``batch'' and ``day'' interchangeably, as it is more intuitive to think
in terms of a temporal framework. 
Note that one could use any arbitrary time period, \lq\lq day\rq\rq\ is simply an example.
Although we focus on batch updates, our method works equally well with online updates.
%
\subsection{Scenarios Description} \label{sec:scenarios}
We consider three scenarios that only differ 
on what happens at the end of a pre-specified time step $t$. 
At the start of the experiment, we initialize the three models using a standard normal prior
$\mathcal{N}(0,1)$. 
At the end of each day, the models are trained in batch with the day's observed data.
At the end of day $t$, the following three scenarios are applied:
\begin{description}
 \item[\textbf{BLIP:}] Our base model, which stands for \textbf{B}ayesian \textbf{LI}near \textbf{P}robit as mentioned earlier. We update the model in batch with day $t$ data. This is the Bayesian regression method that our model builds and improves upon, and the basis of our production GLM TS bandit in Section~\ref{sec:liveExpAll}. One can use any Bayesian GLM model, we opt for BLIP as a fast and scalable state-of-the-art production system \citep{Blip14Facebook, Ding19WholePage}.  
 \item[\textbf{BLIPBayes:}] We reset the model. We bootstrap all the data observed until day $t$ 
   (as per Section~\ref{sec:Bootstrapping})
   and train the model on the bootstrapped data. We then use the Empirical Bayes 
   computation of Section~\ref{EB} to compute separate informative priors $\hat{\tau}^2_k$ 
   for the first and second order features. We then restart the model with 
   a $\mathcal{N}(0,\hat{\tau}_1^2)$ prior for the first-order features and a $\mathcal{N}(0,\hat{\tau}_2^2)$
   prior for the second-order features. We update the new EB model with the original data observed
   up until day $t$. This is our proposed model.
 \item[\textbf{BLIPTwice:}] We update the model twice, first with the 
   same bootstrapped data as BLIPBayes and second with all the data observed until day $t$.    
   The rational is that BLIPTwice uses the same amount of data as used by BLIPBayes. BLIPTwice is
   not a proper optimization algorithm, we only include it when studying the effect of data reuse.
\end{description}

We train the three models on the same batches. At the end of day $t$, 
adaptive lasso
prunes the feature space identically for the three scenarios. 
We start evaluating the models after the day $t$ update. 
At the end of each batch,
we evaluate the models on 
the holdout testing set using binary log loss (cross-entropy).
Let $n$ be the number of observed data points, $y_j \in \{0,1\}$ 
the true binary label of instance $j$, and $P_j$ the model's predicted 
probability of $y_j = 1$, then:
\begin{equation}\label{logloss}
  \text{Log Loss} = -\frac{1}{n}\sum_{j=1}^{n}\ [y_j \ln P_j+(1-y_j)\ln (1-P_j)].
\end{equation}
Note that the output variable in BLIP model given in 
Equation~(\ref{eq:BLIP}) takes values in $\{-1,1\}$ set. We adjust 
our labels before computing the log loss given in Equation~(\ref{logloss}). 

\subsection{Simulation Results} \label{fourExpr}
We now provide the results of our experiments using 
the Adult dataset. Specifically, we study the effect of first-order features
and batch size across our three scenarios, BLIP, BLIPBayes, and 
BLIPTwice. We also study the effect of prior reset time and prior variance on 
BLIPBayes, and compare them against BLIP. 
All these experiments used the same total traffic amount for 
training: 30,162 records from the Adult dataset.

\subsubsection{First-Order Feature Effect:}
Based on the findings of Section~\ref{sec:dataPreprocessing}, and after day $t=1$, we compute
the hierarchical empirical prior over a bootstrapped data pruned using
adaptive lasso. Adaptive lasso retained 7 first-order and 11 second-order features.
As we surmise that the first-order features may hold more predictive 
power during the first batches, we test retaining all 13 first-order features alongside the
11 pruned second-order features.
Keeping all first-order features results in 
$\hat{\tau}_1^2=0.852$ and $\hat{\tau}_2^2=0.241$. Keeping only selected
first-order features returns $\hat{\tau}_1^2=0.714$ and $\hat{\tau}_2^2=0.460$.
\begin{figure}[!ht]
    \centering
       \subfloat[]{\includegraphics[width=.33\columnwidth]{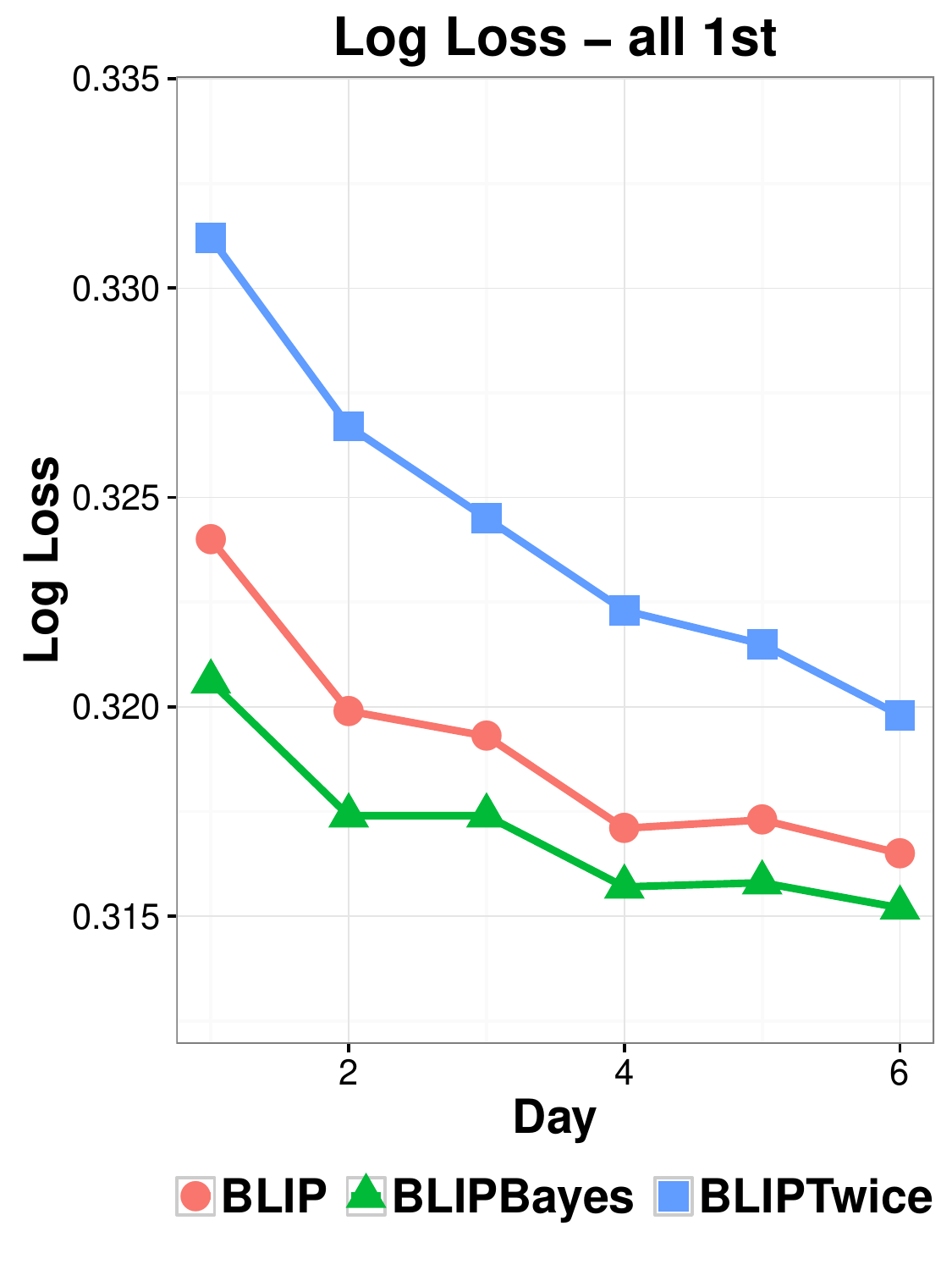}}
       \subfloat[]{\includegraphics[width=.33\columnwidth]{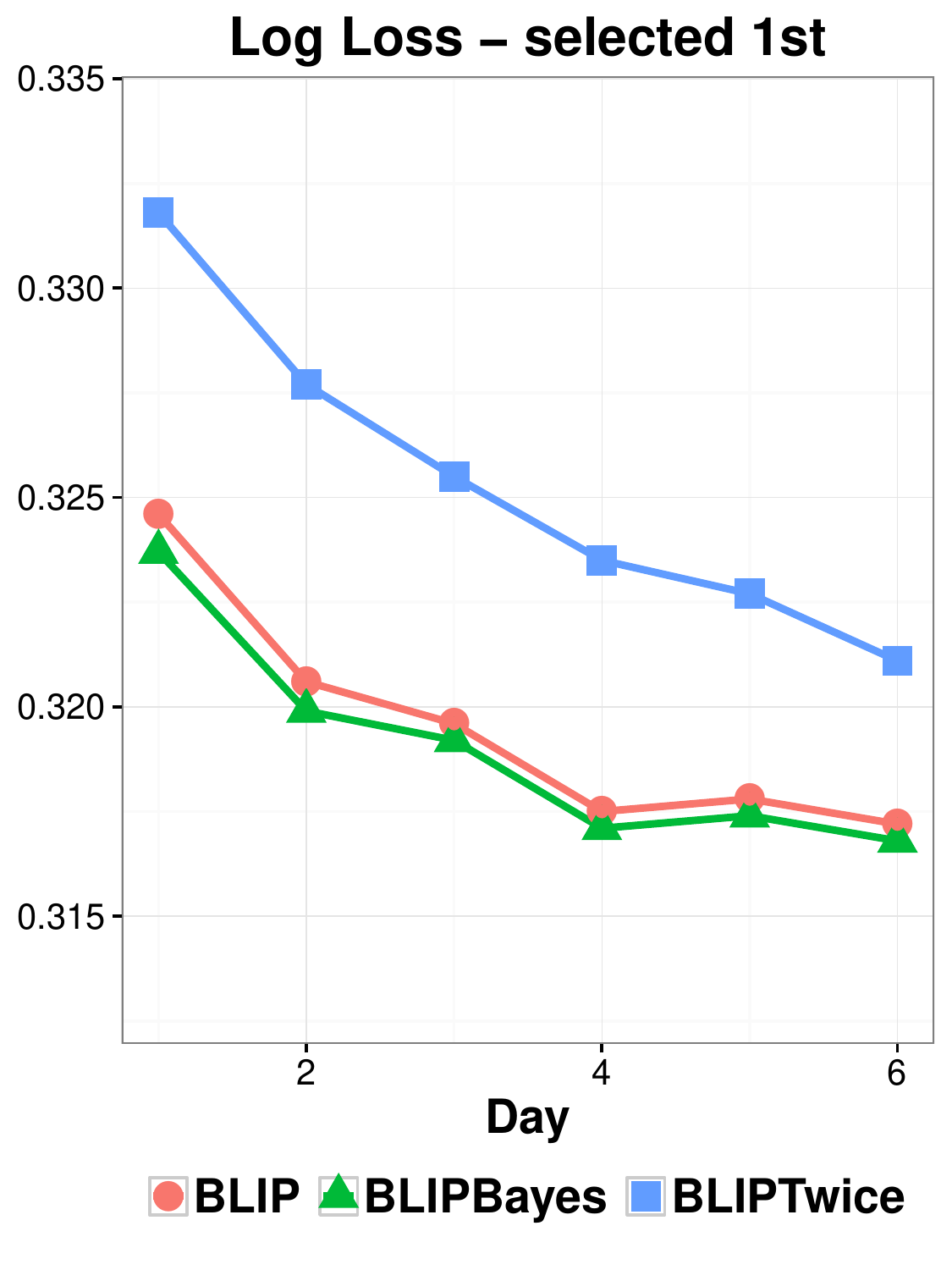}}
		\caption{Log loss with prior reset after day 1. 
             Figure (a) keeps all 13 first-order features and only the 11  
             second-order features selected by adaptive lasso. 
             Figure (b) only keeps the 7 first-order and 11 second-order 
						 features selected by adaptive lasso.}
    \label{lossComparison_40k_merged1And4}
\end{figure}

Figure~\ref{lossComparison_40k_merged1And4} plots the log loss of our scenarios. 
We observe that BLIPBayes outperforms BLIP and BLIPTwice in both cases.
We suspect that the bad performance of BLIPTwice is due to overfitting to the first $t$ batches.
Retaining all first-order features improves the three methods' prediction accuracies.
We also note that keeping all first-order features results in a markedly better performance
for BLIPBayes. This may be due to a better estimate of $\hat{\tau}_1^2$.
In our subsequent experiments, we retain all first-order features, and use
adaptive lasso to prune the second-order features.
%
\subsubsection{Effect of Prior Reset Time:} \label{BayesReset3rdDay}
In this experiment, we reset the empirical prior at the end of day $t=3$, 
after observing 15,000 samples. Adaptive lasso
pruning retains 11 second-order features. 
EB results in $\hat{\tau}_1^2=0.862$ and $\hat{\tau}_2^2=0.414$.

\begin{figure}[!ht]
  \centering
  \includegraphics[width=0.57\columnwidth]{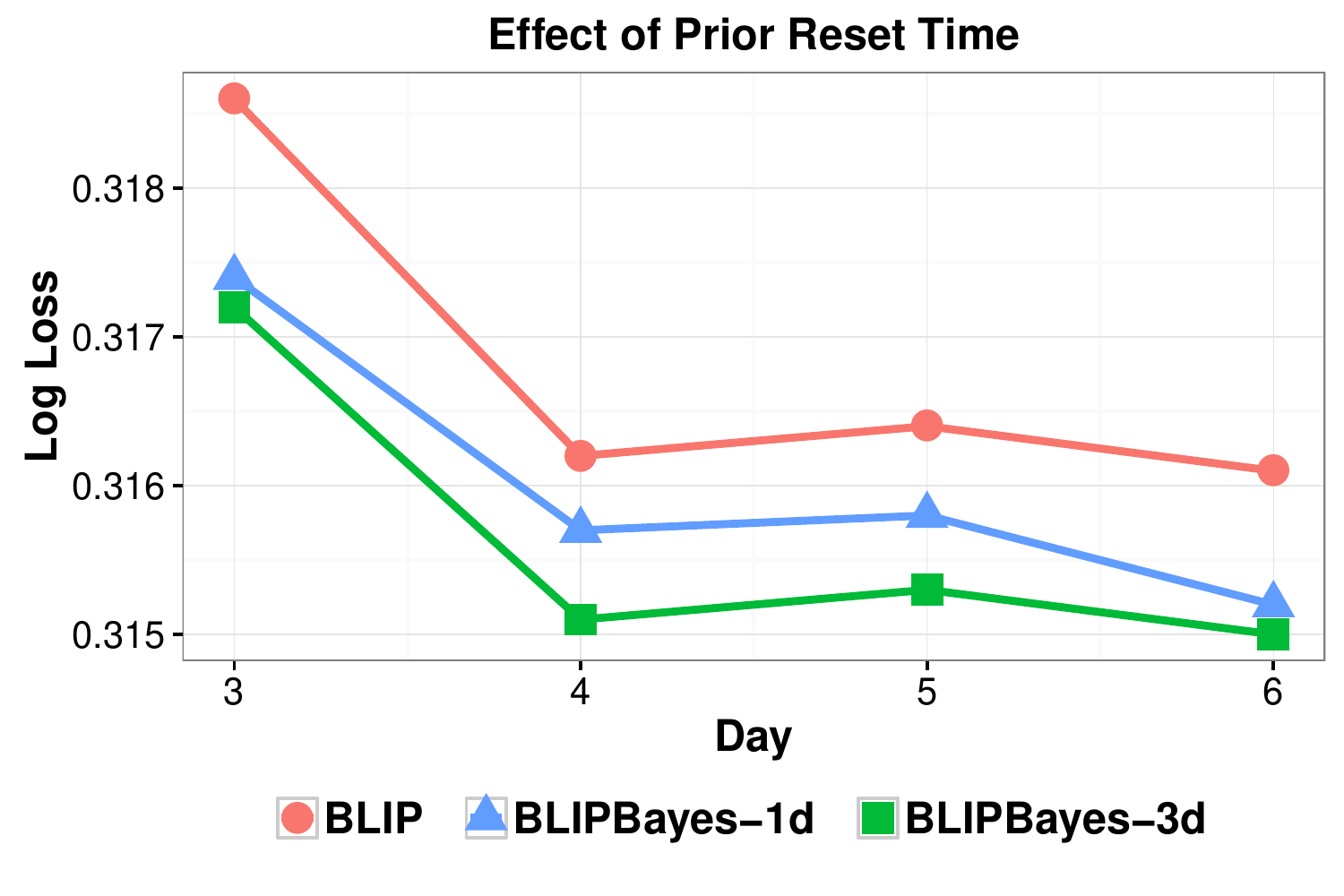}
  \caption{Comparing log loss of no prior reset (BLIP), prior reset after day 1 (BLIPBayes-1d), and 
  after day 3 (BLIPBayes-3d). 
    Using more data to compute the empirical prior improves performance.
  }
  \label{priorReset}
\end{figure}

Figure~\ref{priorReset} compares the log loss for BLIP, BLIPBayes with $t=1$ reset, and 
 BLIPBayes with $t=3$ reset. BLIPBayes gives lower log loss values and therefore
higher prediction accuracy when using more data for constructing the prior. 
As we observed when comparing all versus selected first-order features, 
more data improves performance.

%
\subsubsection{Small Batch Dataset:} \label{30DaysRuns}
%
In this experiment, we divide the train dataset into thirty batches, 
each with 1000 data points. Our objective is 
to observe the performance of EB when training 
occurs on a longer period with smaller data batches.  
We reset the prior at $t=1$, but this time we bootstrap and train over 12,000 instances.
Adaptive lasso retains 21 second-order features. 
EB results in $\hat{\tau}_1^2=0.799$ and $\hat{\tau}_2^2=0.132$. 

\begin{figure}[!ht]
  \centering 
	   \includegraphics[width=.7\columnwidth]{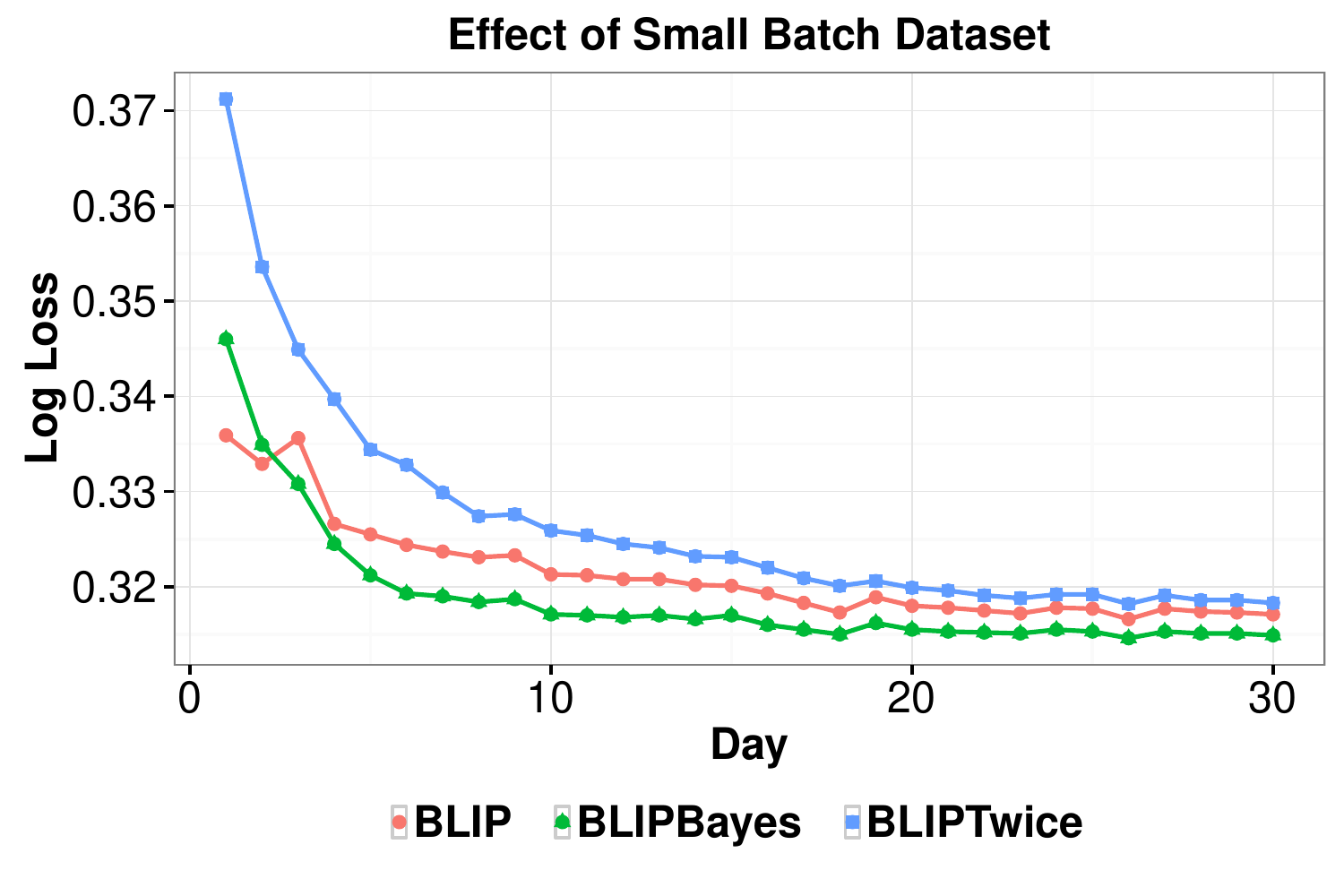}
	\caption{Log loss over smaller batch dataset. The total traffic amount is unchanged.}
	 \label{logloss_30Days}
\end{figure}

Figure~\ref{logloss_30Days} plots the log loss over the thirty days.
BLIPBayes outperforms both other methods.
What is remarkable is that the BLIPBayes advantage persists over the whole range,
indicating that such a method can be especially valuable for small batch training.

%
\subsubsection{Effect of Prior Variance $\tau^2_k$:} \label{extremeRuns}
%

It is possible that our improvements simply stem from the fact that our empirical prior
variance is below 1, its non-informative counterpart. 
This would not adhere to the finding uncovered in Section~\ref{sec:regret},
where $regretBound$ increases as $\tau_k$ shrinks.
To shed additional light on this issue, we simulate the worse-case scenario
of $\hat{\tau}_1 = \hat{\tau}_2$, and vary their values from $\approx 0^+$ onward.
Recall from our first experiment setting that EB returns 
$\hat{\tau}_1^2=0.852$ and $\hat{\tau}_2^2=0.241$ by following the formula
given in Equation (\ref{eq:BayesVariance}).
We experiment with additional $\hat{\tau}^2_k$ settings, namely
 $\hat{\tau}_1^2=\hat{\tau}_2^2=5$, $\hat{\tau}_1^2=\hat{\tau}_2^2=0.1$, 
and $\hat{\tau}_1^2=\hat{\tau}_2^2=0.01$. 

\begin{figure}[!ht]
  \centering 
  \includegraphics[width=0.63\columnwidth]{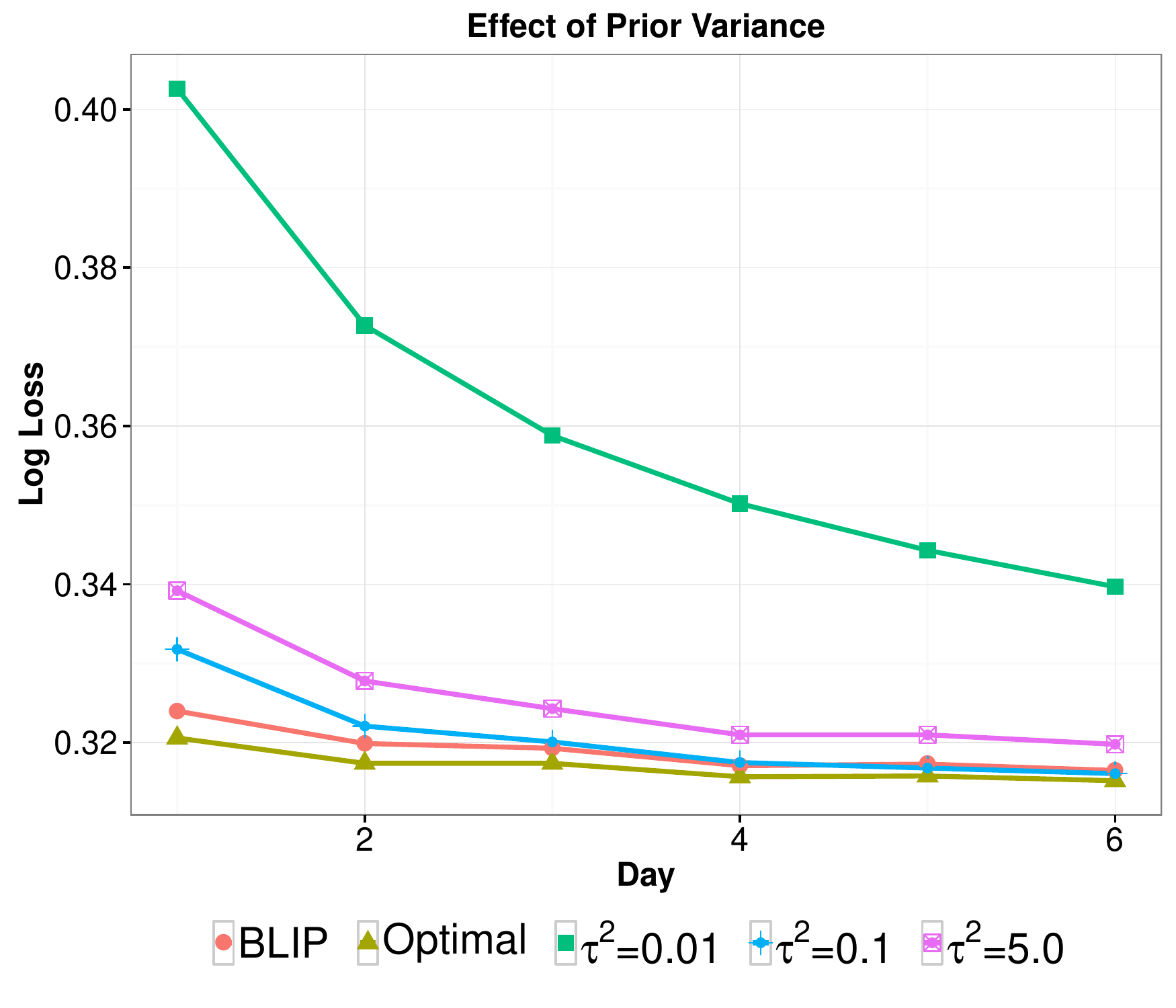}
  \caption{Comparing the effect of $\tau^2_k$ on performance.
    We varied $\hat{\tau}^2_k$ away from its optimal EB values, 
    $\hat{\tau}_1^2=0.852$ and $\hat{\tau}_2^2=0.241$.
  }
  \label{logloss_all}
\end{figure}

Figure~\ref{logloss_all} plots the log loss 
for the aforementioned scenarios, alongside the EB values (``optimal'') and BLIP.
We observe that the optimal BLIPBayes consistently 
outperforms all other variations. 
We also note that BLIP
(with its $\hat{\tau}_1^2=\hat{\tau}_2^2=1$ priors) outperforms the non-optimal BLIPBayes versions
(with a negligible overlap with $\hat{\tau}^2_k=0.1$ after day 5). 
This suggests that one needs to set the hierarchical priors in a principled manner,
and that $\hat{\tau}^2_k < 1$ is not necessarily better.
We also note that $\hat{\tau}^2_k=0.01$ achieves the worst results by far, concordant with our Section~\ref{sec:regret}
finding, and that $\hat{\tau}^2_k=5$ under-performs $\hat{\tau}^2_k=0.1$. 
This suggests that erring towards a large prior variance is more easily overcome by data 
than erring towards a small prior variance \citep{BaBa20}. In fact, in a Bayesian setting,
the relative effect of the observed data on the posterior increases with a larger 
prior variance.

\section{MAB Live Experiments}\label{sec:liveExpAll}
We now examine the performance of EB on the live production system 
described in \citet{hill2017kdd}. 
%

\subsection{Experimental Settings}\label{sec:liveExp}

The problem we address is the selection of a layout $A$ of a web page with the objective of 
maximizing the expected value of a binary reward $y \in \{-1,1\}$. 
Each layout $A$ is generated from a common template that contains $D$ widgets representing the contents of the page. The $i^{th}$ widget has $n_i$ alternative variations for its content; however, for simplicity of notation, we assume equal number of variations per widget 
denoted by $n$. A web page has thus $n^D$ possible layouts. 
We represent a layout as $A \in \{1,2,\ldots,n\}^D$, a $D$-dimensional vector where 
$A[i]$ denotes the content chosen for the $i^{th}$ widget. 

We model this scenario as a stochastic contextual multi-armed bandit with linear reward. 
Our model is a TS generalized linear MAB with a probit link 
function and $\mathcal{N}(0,1)$ prior~\citep{kveton2020randomized}. Its core is
the same model used for classification in Section~\ref{sec:model}.
At each time step $t$, we select the layout to present to a user using Thompson Sampling.
We sample the model parameters from their posterior and pick the layout that maximizes the reward,
following Equation~(\ref{eq:TS}).

Let $B_A$ be the featured representation of a given layout.
Let $R = +1$ indicate the user took the desired action which refers to the 
service purchase in our live experiment 
and $R = -1$ indicate otherwise. 
The reward probability is modeled as:
\begin{equation}
p_r(R|A) = \Phi(R * B_{A}^\top W),
\end{equation}
where $W$ presents the unknown parameter vector and 
$\Phi$ refers to the CDF of the standard Normal distribution. Note that 
this is the same model as in Equation (\ref{eq:BLIP}). 
Although the original formulations take user context into account, this work
investigates the case where the features only reflect the layout content.
Furthermore, we only consider pair-wise interactions and assume that 
the effects of higher order interaction terms are negligible. Therefore, our linear model becomes:
\begin{equation}
B_A^\top W  = W^0 + \sum_{i=1}^{D}W^1_i(A) + \sum_{j=1}^{D}\sum_{k=j+1}^{D}W^2_{j,k}(A), 
\end{equation}
where $W^0$ is a common bias weight, $W^1_i(A)$ is a weight associated with the content in 
the $i^{th}$ widget, and $W^2_{j,k}(A)$ is a weight for the interaction between the contents of the $j^{th}$ 
and $k^{th}$ widgets. For EB computations, $W^1$ are the weights of the first-order features, and $W^2$ the weights of 
the second-order features.

At each time step $t$, we select the layout to present to a user using Thompson Sampling.
A layout is shown proportional to its probability of being optimal.
We sample the model parameters from their posterior and pick the layout that maximizes the reward.
Let $\tilde{W_t}$ be the sampled weights.
We present the layout that solves the following optimization problem:
\begin{equation}\label{eq:TS_MVT}
 A_t=\argmax_{A} \ B_A^\top\tilde{W_t}.
\end{equation}
As we have $N^D$ layouts, a standard MAB results in $N^D$ arms. 
For small $N$ and $D$ values, we solve this optimization using exhaustive search.
This is the case in our live experiments in Section~\ref{sec:liveResults}.
In settings with large $N$ and $D$ values, we recommend deploying a
greedy hill climbing optimization procedure with random restarts (\cite{CaBe02}).

In our live experiments, we aim at optimizing a message that promotes the purchase of an Amazon service.
The message had $D=4$ widgets with $n_i = \{2,3\}$ possible options per widget, for a total of
$d = 24$ distinct combinatorial layouts. 
The target is binary, whether the customer purchased 
the service or not. The messages were shown to the selected customers 
during a browsing session on Amazon.com.

We performed A/B tests with three treatments, a production baseline algorithm, the 
standard probit bandit, and EB applied to the probit bandit. 
We randomly diverted a constant subset of our 
traffic to this experiment, with the standard and EB bandits receiving equal shares.
The baseline algorithm has a different pool of messages, 
and dynamically adjusts to seasonal shifts in an adversarial manner. 
In order to study the effect of the empirical prior reset, we disable
seasonality adjustment for both standard and EB bandits, deploying them as stochastic MABs. 

\subsection{Live Results}\label{sec:liveResults}

We start both bandits with a 
random phase, where the bandits allocate traffic equally between their 24 layouts.
We exclude this random phase from our plots and analysis.
We then compute the empirical prior and re-train the EB MAB on the random data. 
Unlike the simulation experiments, we do 
not prune features nor do we bootstrap the data for EB computation.
As we do not know the
ground truth, we can not compute regret nor log loss. Instead, we compute the
cumulative success rates of each bandit.

\subsubsection{Traffic Effect}
\begin{figure}[!ht]
  \centering 
  \includegraphics[width=0.7\columnwidth]{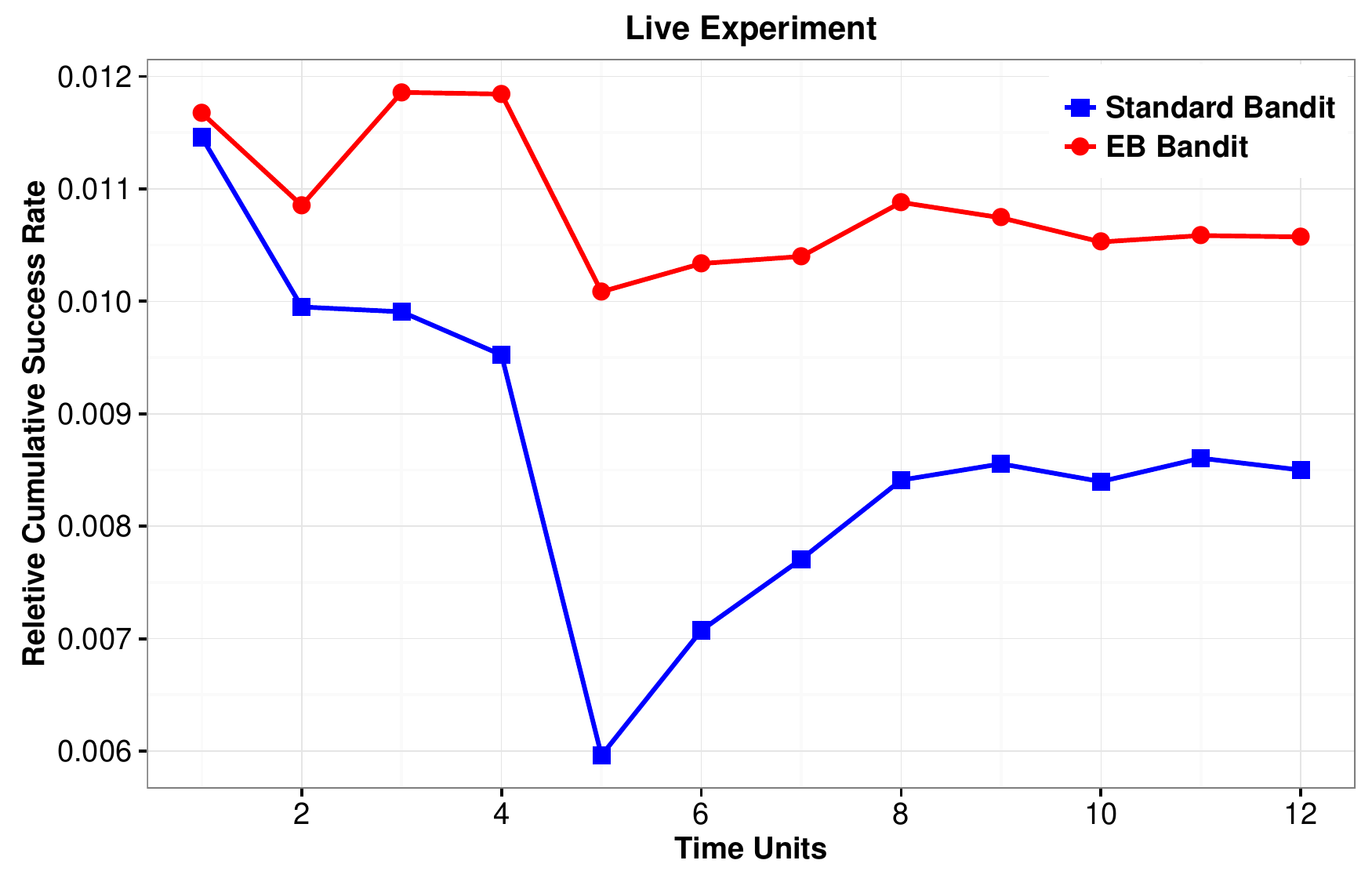}
  \caption{Cumulative success rate in first live experiment, relative to baseline final 
    cumulative success rate.}
  \label{fig:prodExp}
\end{figure}

In the first experiment, we diverted a constant traffic percentage to our MABs.
We set the random phase to 3 time units, after which we ran the experiment for 12 time units during a non-seasonal time frame.
EB resulted in $\hat{\tau}^2_1 = 0.613$ and $\hat{\tau}^2_2 = 0.195$. Figure~\ref{fig:prodExp} plots the cumulative 
success rate of each bandit, relative to the baseline's final cumulative success rate. 

The EB MAB clearly dominates the standard MAB in cumulative rewards. We notice that EB MAB
stabilizes after five time units, while the standard MAB needs eight to plateau. This is an indication
that EB MAB converged faster. After $t=8$, both bandits attain the same performance, maintaining 
similar success rates. From $t=1$, the empirical prior focused the exploration of the EB bandit, while the standard bandit had to fully explore the space of suboptimal layouts. This resulted in improved cumulative gain, as reflected by the gap in the relative cumulative success rates. 

The dip at $t=5$ is due to a production update delay that affected all algorithms, including the baseline. The  standard bandit suffers the most, as it was still exploring and deployed a suboptimal policy. EB proves robust to such disruptions, a crucial quality as data issues are commonly observed in any production system. 

We followed this experiment with a shorter one (4 time units total) during a fixed season, 
where we reduced the traffic by a quarter, 
and reduced the random phase to 1 time unit. 
EB resulted in $\hat{\tau}^2_1 = 0.489$ and $\hat{\tau}^2_2 = 0.087$.
At the end of the experiment, 
we compared the cumulative performance of both MABs against the baseline. 
Using two-tailed proportion z-test with pooled variance, 
EB bandit had a $0.05$ p-value, and standard 
bandit a $0.11$ p-value. EB MAB significantly outperformed the baseline, 
and effectively converged, while the
standard MAB did not.
Another indication that EB can be most valuable for smaller traffic cases.
We report significance against the baseline to measure convergence. Testing EB MAB versus standard MAB is not significant and is irrelevant for us, as both converge to the same solution given enough data.

\subsubsection{Seasonality Effect}\label{sec:seasonalityEffect}
%
To test the impact of seasonality, we ran our final experiment over a holiday season, 
with significant pre and post seasonal changes. The experiment extended over 54 time units, following a 
random phase of 2 time units, and did not suffer from major production system delays.
We dialed up the traffic assigned to each MAB, 
doubling the traffic on average, while tripling it in the first two weeks.
EB resulted in $\hat{\tau}^2_1 = 0.448$ and $\hat{\tau}^2_2 = 0.133$.

\begin{figure}
  \centering 
  \includegraphics[width=0.75\columnwidth]{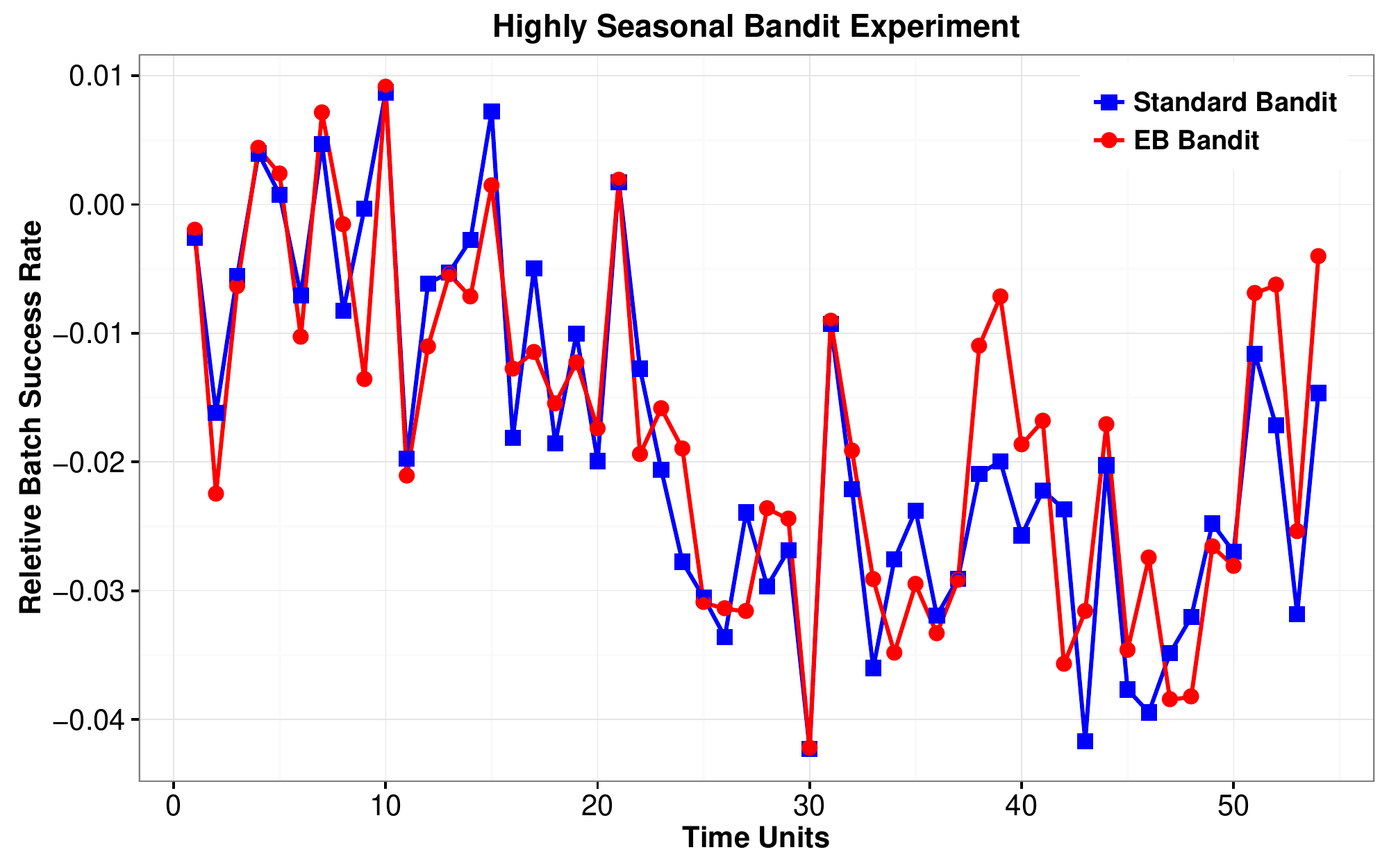}
  \caption{Batch success rate in highly-seasonal third live experiment, relative to the baseline success rate.}
  \label{fig:exp3}
\end{figure}

At this level of high traffic, both bandits behaved indistinguishably. The empirical prior had little to no effect, as expected in higher traffic situations. As the two bandits' cumulative success rates were almost identical,
Figure~\ref{fig:exp3} plots their batched (per unit time) success rates. One can see how they fluctuate similarly.
We also note that seasonality had an adverse effect on the bandits, as both 
lagged behind the seasonality-aware baseline. Recall that we had disabled seasonality adjustments for both bandits, but not for the baseline. This explains their negative relative success rates compared to the baseline.

\section{Discussion and Conclusion} \label{Discussion}
We conclude by analyzing and discussing our findings.

\subsection{Discussion}
%

It is interesting that, empirically, most $\hat{\tau}^2_k$ priors computed during our experiments had values below $1$.
Recall from Section~\ref{sec:regret} that $\tau^2_k$ affects the bandit regret bound through constants
$p$ and $c$. Parametrizing $p, c$ using $\tau^2_k$ in Equations~(\ref{eq:regretGLM}) and~(\ref{eq:regretConst}), we notice that $regretBound(\tau^2_k < 1) > regretBound(\tau^2_k = 1)$. Here $regretBound$ refers to the r.h.s of Equation~(\ref{eq:regretGLM}). As a small $\tau^2_k$ reduces exploration, this could explain the higher regret bound~\citep{LiLi16}. But in our experiments, we observe that EB constantly outperforms the standard bandit. This suggests that imposing an EB hierarchical prior to decouple the learning rates of the first and second-order features could have a tangible effect on the regret.

In a supervised learning setting, the unbiasedness and consistency results of Section~\ref{sec:theory}, coupled with our experimental results, suggest that our EB computation method is likely to yield a lower log loss. In a bandit setting, 
we do observe an empirical improvement, even when $regretBound$ may be interpreted as suggesting otherwise for $\tau^2_k < 1$. This seems to indicate a trade-off between loosing on exploration potential, and gaining by introducing a meta-prior that decouples learning rates. In cases where there is a difference between the effects of different categories' features, i.e. when true $\tau_1 \neq \tau_2 \neq 1$, EB would find a better prior that is likely to improve on the regret. Since the choice of the hierarchical prior variance has no effect on the asymptotic regret $\tilde{O}(d^{3/2} \sqrt{T})$, one may loose little by applying the EB method, even if true $\tau_1 = \tau_2 = 1$.

In all of our simulation and live experiments, we always had $\hat{\tau}_1^2 > \hat{\tau}_2^2$, 
reflecting a difference in the effects of the first and second order features.
This confirms our initial conjecture, that the second-order features are likely to be
less important. This result is likely to generalize to many applications.

By grouping the first-order and second-order features together and imposing a hierarchical prior, we 
effectively clamped each category's weights together.
Since the relative effect of the observed data on the posterior increases with a larger prior variance,
our method is putting more weight on the first-order features, and is shrinking the second-order effect. 
The model thus focuses on learning the first-order effects first. This may explain the 
increased stability and convergence speed of the EB model.

Of interest is how the improvement is correlated with the amount of available data. Our findings
suggest that the EB improvement is most marked in cases of low to medium traffic, and is lost at high 
traffic. This is promising, as low traffic cases are the hardest to optimize. At very low traffic, 
direct computation of empirical prior variances may fail, with $\hat{\tau}^2_k < 0$. 
Even if $\hat{\tau}^2_k \approx 0^+$, one needs guardrails to preserve a sublinear regret bound, as per 
Section~\ref{sec:regret}. One may either wait longer
before computing the prior (see Section~\ref{BayesReset3rdDay}), 
bootstrap from available data (see Section~\ref{sec:Bootstrapping}), prune the feature space 
(see Section~\ref{sec:dataPreprocessing}), add a minimal $\tau_k$ threshold, or perform transfer learning. We leave investigating the last two suggestions for future work. 

In order to study the effect of the empirical prior reset, we disabled
seasonality adjustment for both standard and EB bandits in Section~\ref{sec:liveExpAll}. 
For short-period experiments, it is a reasonable approximation to assume stationarity. We do not recommend 
disabling seasonality adjustment for long-period experiments, as indicated in Section~\ref{sec:seasonalityEffect}.
Since we compute $\hat{\tau}^2_k$ close to the start of the experiment, the effects of seasonality and algorithmic adjustments have little bearing on its value. We confirmed this finding in subsequent experiments. 
For long-period experiments where seasonality adjustment is paramount, the choice of the prior becomes less relevant with additional data.

As with most Bayesian techniques, we model the meta-prior using the conjugate distribution of the posterior. As we used a probit regression, the feature weights are modeled as Gaussians, with a Gaussian conjugate prior. This work seeks to improve the starting prior of an existing model, without questioning the validity of the prior's distribution family. If one has no valid distribution shape for the prior, techniques like the ones explored in \citet{luedtke2020adversarial} could help. Even if the prior distribution is unknown or misspecified, one could invoke the Central Limit Theorem to justify the choice of a Gaussian variance representation for Equation~(\ref{eq:BayesVariance}).

\subsection{Future Work}
%

Our findings raise multiple questions and opportunities for future work.
In this paper, we decouple learning rates by imposing a hierarchical meta-prior, and estimate its variance prior using empirical Bayes. One could investigate other methods to estimate the priors, and compare techniques to ensure non-degenerative empirical priors.
This work specifically focuses on first and second-order features, with different incidence rates. 
One could further investigate the effect of incidence rates on the meta-prior computation. 

We used EB to effectively bootstrap an existing model using its own data. It could be possible to apply EB to 
different but related use-cases as a transfer learning technique \citep{BaSi19}. This may be valuable if the related use-case has a large data volume, while the target scenario is highly sparse.
Our technique requires pre-specified feature categories, a harder challenge is to automatically infer the feature categories.

Separately, decoupling learning rates is a valuable by-product of our method. It can be attributed to:
(i) the features are grouped in different categories and share different priors, and (ii) the EB approach provides more accurate per-category prior estimation. One can further study the dynamics of decoupling learning rates using EB, as well as comparing it to other methods.

Another potential research question is whether our meta-prior framework can be extended to dynamic Bayesian learning in general settings. One such application is dynamic learning with optimal stopping, which occurs in launch/innovation and investment strategies. As one example, optimal policies in \cite{HaSu15} are characterized based on optimal lower and upper critical beliefs that depend upon posterior estimations. A more informative prior leads to more accurate estimations of these parameters at any given time. This, in turn, can increase or decrease the length of the learning continuation region described in \cite{HaSu15}. Data deficiency issues may also arise in optimal stopping problems due to the potential shorter learning horizon. Leveraging existing structure given the application domain and problem formulation may benefit meta-prior learning and help overcome potential data deficiencies.

\subsection{Conclusion} \label{Conclusion}

This study proposes an Empirical Bayes method that leverages early observed data in a given experiment to compute category-specific informative meta-priors for that experiment.
Such informative meta-priors have the added effect of decoupling learning rates of different categories of features.
We show that our GLM-based estimator is unbiased, and prove its strong consistency in the regime where the feature estimates are independent. Furthermore, we show that our EB BLIP-based TS bandit algorithm retains an optimal $\tilde{O}(d^{3/2} \sqrt{T})$ regret bound.
We apply this method to decouple the learning rates of first and second order features in a supervised learning and a bandit setting.
Our empirical results reveal that our Empirical Bayes technique leads to higher cumulative rewards and lower convergence time for bandits and improves prediction accuracy for classifiers.
Of special note are the observed improvements in cases of smaller traffic, leading us to believe that
Empirical Bayes may offer an adequate solution for the challenges of sparse-data 
optimization.

\section*{Acknowledgments} \label{Acknowledgments}
The authors thank Gregory Duncan, Sham Kakade, Milos Curcic, Lalit Jain, Omid Rafieian, 
Devin Didericksen, Yi Liu, Karthik Mohan, Emily Ikeda-Flowers, Calvin Kwok, Zachary Austin,
and Jason Yang for their support and helpful discussions. 

\bibliographystyle{informs2014} 
\bibliography{paper-bibliography} 

\newpage

\begin{APPENDIX}{Computation of Bandit Cumulative Regret Constants}
Here, we provide details on how to compute
anti-concentration probability $p$ and constant values
$c$ and $c'$ in the regret upper bound given in 
Equation~(\ref{eq:regretGLM}) and Corollary \ref{thm:corollary4}.

\citet{AbLa17} show that TS can be defined as a
randomized algorithm that is built upon a Regularized Least 
Square (RLS) estimate, given past history of arms pulled and 
rewards observed. They show that TS samples from a perturbed 
parameter that depends on this RLS estimate and a multivariate 
distribution that follows anti-concentration and concentration 
properties. 
We first restate these properties and compute their corresponding 
parameter values for our EB BLIP TS model.

\begin{definition}\label{def:concentration} (As provided in \citealp{AbLa17}).
$D^{TS}$ is a multivariate distribution on $\mathbb{R}^d$
absolutely continuous with respect to Lebesgue measure which
satisfies the following properties: 
	\begin{enumerate}
	\item \emph{Anti-concentration property:} There exists a
	strictly positive probability $p$ such that for any $u \in \mathbb{R}^d$ with $||u||=1$, 
	\begin{equation}
 	P_{\eta \sim D^{TS}} \Big( u^T\eta \ge 1 \Big) \ge p.
	\end{equation}
	\item \emph{Concentration property:} There exists $c, c'$ positive constants such that $\forall \delta \in (0,1)$,
	\begin{equation}\label{const:c}
 	P_{\eta \sim D^{TS}} \Bigg( ||\eta|| \le \sqrt{cd \log \frac{c'd}{\delta}} \Bigg) \ge 1-\delta.
	\end{equation}
	\end{enumerate}
\end{definition}

In our EB BLIP, $\eta \sim \mathcal{N}(0, \textrm{I}_{d,\tau})$
where $\textrm{I}_{d,\tau}$ denotes the diagonal covariance matrix of size $d*d$ with 
diagonal values $\tau_i^2,\ \forall i=1,...,d$. Let $\eta_{\tau_i}$ denote the 
i-th component of $\eta$. Hence, $\eta_{\tau_i} \sim \mathcal{N}(0, \tau_i^2),\ \forall i=1,...,d$.
We have 
\begin{equation}\label{p:regretConst}
P\Big( u^T\eta \ge 1 \Big) \ \ge \ P\Big( \eta_{\tau_{\min}} \ge 1 \Big) 
                             \ = \ \frac{1}{2}\ \textrm{erfc} \Bigg( \frac{1}{\tau_{\min}\sqrt{2}} \Bigg) 
														 \ \ge \ \frac{e^{\frac{1}{2}-\frac{1}{\tau_{\min}^2}}}{4 \sqrt{\pi}},
\end{equation}
where $\tau_{\min}=\min_k \tau_k$ and $\eta_{\tau_{\min}} \sim \mathcal{N}(0,\tau_{\min}^2)$. 
The first inequality holds since the anti-concentration property must be satisfied 
for any arbitrary $u \in \mathbb{R}^d$. The second equality and third inequality 
are due to the definition of the Gaussian Q-function and its lower bound, given in \citet{ChCo11}. 
Let $Q(x)$ denote the Gaussian Q-function for $X \sim \mathcal{N}(0,1)$,
\begin{equation}
Q(x):=P(X>x)=\frac{1}{2}\ \textrm{erfc} \Bigg( \frac{x}{\sqrt{2}} \Bigg) \quad \textrm{where} \quad 
            \textrm{erfc}(x)=\frac{2}{\sqrt{\pi}} \int_{x}^{\infty} e^{-t^2} dt, x>0.
\end{equation}
\cite{ChCo11} prove that $f(x)=\alpha e^{-\beta x^2}$ is a lower bound on $Q(x)$
for values of $\beta>1$ and $0< \alpha \le \sqrt{\frac{2e}{\pi}}\frac{\sqrt{\beta-1}}{\beta}$. 
We set $\beta=2$ and $\alpha=\frac{1}{2}\sqrt{\frac{2e}{\pi}}$ to obtain the 
third inequality in (\ref{p:regretConst}). Hence, by definition~\ref{def:concentration},
we have anti-concentration probability $p=\frac{e^{\frac{1}{2}-\frac{1}{\tau_{\min}^2}}}{4 \sqrt{\pi}}$.

We now provide details on how to compute concentration constants 
$c$ and $c'$ in Equation~(\ref{const:c}). For any $\gamma>0$, and using the complement rule and union bound,
we have:
\begin{equation}
P \left(||\eta|| \le \gamma \sqrt{d}\right) 
                     \ge P \left( \forall \ 1 \le i \le d, \ |\eta_{\tau_i}| \le \gamma \right)
										 = 1 - P \left(\exists i \ \ s.t.\ \ |\eta_{\tau_i}| > \gamma \right)
									  \ge 1-\sum_{i=1}^{d} P\left( |\eta_{\tau_i}| > \gamma \right).
\end{equation}
Let $\tau_{\max}=\max_k \tau_k$ and $\eta_{\tau_{\max}} \sim \mathcal{N}(0,\tau_{\max}^2)$.
We have, 
\begin{equation}
 P \left(||\eta|| \le \gamma \sqrt{d}\right) \ge 1-d P \left( |\eta_{\tau_{\max}}| > \gamma \right)
                                          \ge 1-2de^{-\frac{\gamma^2}{2\tau_{\max}^2}},
\end{equation}
where the last inequality follows from applying the Chernoff bound (standard Gaussian concentration inequality). 
Using Equation~(\ref{const:c}), we substitute for 
$\delta=2de^{-\frac{\gamma^2}{2\tau_{\max}^2}}$,
which results in $\gamma=\sqrt{2\tau_{\max}^2 \log \frac{2d}{\delta}}$,
$c=2\tau_{\max}^2$, and $c'=2$.

\end{APPENDIX}

\end{document}